\documentclass[aps,aplml,reprint,onecolumn,groupedaddress]{revtex4-2}

\usepackage[margin=1in]{geometry}
\usepackage{hyperref}  
\usepackage{setspace}
\usepackage{graphicx}
\usepackage{subcaption}
\usepackage{amsmath}
\usepackage{makecell}
\usepackage{tabularx}

\begin{document}


\title{Assessing Language Models for Scientific Applications: Inertial Confinement Fusion as a Use Case}

\author{P. B. Radha}
\email[]{radha@lanl.gov}
\affiliation{Los Alamos National Laboratory, Los Alamos, NM}
\author{D. D. Meyerhofer}
\affiliation{Los Alamos National Laboratory, Los Alamos, NM}
\author{M. Bhattarai}
\affiliation{Los Alamos National Laboratory, Los Alamos, NM}
\author{B. Cheng}
\affiliation{Los Alamos National Laboratory, Los Alamos, NM}
\author{L. Diaz}
\affiliation{Los Alamos National Laboratory, Los Alamos, NM}
\author{C. L. Rousculp}
\affiliation{Los Alamos National Laboratory, Los Alamos, NM}
\author{A. G. Seaton}
\affiliation{Los Alamos National Laboratory, Los Alamos, NM}
\author {G. Gupta}
\affiliation{NVIDIA Corporation, Santa Clara, CA} \author{S. Halverson}
\affiliation{NVIDIA Corporation, Santa Clara, CA}
\author{P. Gurumurthy}
\affiliation{NVIDIA Corporation, Santa Clara, CA}
\author{V. Gutta}
\affiliation{NVIDIA Corporation, Santa Clara, CA}
\author{A. Jhunjhunwala}
\affiliation{NVIDIA Corporation, Santa Clara, CA}
\author{A. Vem}
\affiliation{NVIDIA Corporation, Santa Clara, CA}
\author{P. Cook}
\affiliation{NVIDIA Corporation, Santa Clara, CA}
\author{G. Rowe}
\affiliation{NVIDIA Corporation, Santa Clara, CA}

\date{\today}

\begin{abstract}


Advancing language models for specialized scientific applications requires evaluations of scientific fluency, physical reasoning, and quantitative analyses. Using inertial confinement fusion (ICF) as a test case, we have developed an expert-curated benchmark of 250 questions spanning qualitative reasoning, numerical calculation, symbolic derivation, and equation formulation. Responses have been evaluated using deterministic numerical and symbolic checks together with language-model judges. We compared a variety of out-of-the-box closed-source and open-source large language models with Prospero, a model specialized through combined domain-adaptive pretraining and supervised fine-tuning for ICF. We also conducted initial retrieval-assisted evaluations. Prospero improved its aggregate score by 10.7 percentage points over its base model, with gains across all question types. Frontier models achieved the highest aggregate scores, but critical physics errors persisted, particularly in numerical calculations. We note that domain-specific performance can improve through specialization, but benchmark gains alone do not establish scientific reliability. Retrieval improved performance for a smaller model but yielded no statistically significant improvement for a frontier model. Two scientific stress tests further illustrate useful capabilities and the need for expert assessment of physical assumptions and model validity. Based on these findings, we recommend evaluating existing domain capabilities and retrieval benefits before committing to substantial additional specialization.
\end{abstract}


\maketitle

\section{Introduction}
Super Intelligence (SI) tools are increasingly being developed and deployed for specialized scientific applications, including protein structure prediction \cite{jumper2021alphafold}, materials discovery \cite{merchant2023materials}, experimental control \cite{degrave2022tokamak}, and autonomous experimentation \cite{szymanski2023autonomous}. More recently, large language models (LLMs) have created the possibility of developing general-purpose scientific assistants whose capabilities span multiple tasks. These models can be further specialized through a combination of domain-adaptive pretraining, supervised fine-tuning \cite{chipnemo}, retrieval-augmented generation \cite{lewisRAG, gaoRAG, asaiRAG, edgeRAG}, and reinforcement learning \cite{schulmanRL}. Such specialization is intended to improve familiarity with domain terminology, concepts, and scientific workflows. 

 Standard LLM and scientific reasoning benchmarks - for example, Graduate-Level-Google-Proof Q\&A (GPQA) \cite{GPQA} and Massive Multitask Language Understanding (MMLU) \cite{MMLU} - measure broad knowledge and problem-solving ability, but performance on these benchmarks does not necessarily establish that a model can address the problems encountered by practicing scientists within a specialized field. Domain-specific scientific problems require detailed knowledge, quantitative calculation, symbolic manipulation, identification of the relevant physical processes, selection of an appropriate physical model, and recognition of the assumptions under which that model is valid. A model may therefore perform well on general scientific benchmarks while failing on questions that require the integrated knowledge expected of a domain scientist. 

In hotspot ICF \cite{Nuckolls}, for example, a capsule containing deuterium--tritium (DT) fuel is imploded by a driver such as a high-power laser. During the implosion, kinetic energy of the shell is converted into internal energy of a central hot spot. If sufficiently high temperatures and compression are achieved, DT fusion reactions produce energetic alpha particles that deposit their energy within the compressed fuel. This self-heating can drive a thermal runaway in the fusion reaction rate, leading to ignition. At sufficiently high performance, the fusion energy produced can exceed the energy supplied by the driver, corresponding to target gain greater than unity.

\begin{figure*}[ht!]
\includegraphics[trim=0cm 0cm 0cm 5cm, clip,width=1.0\textwidth]{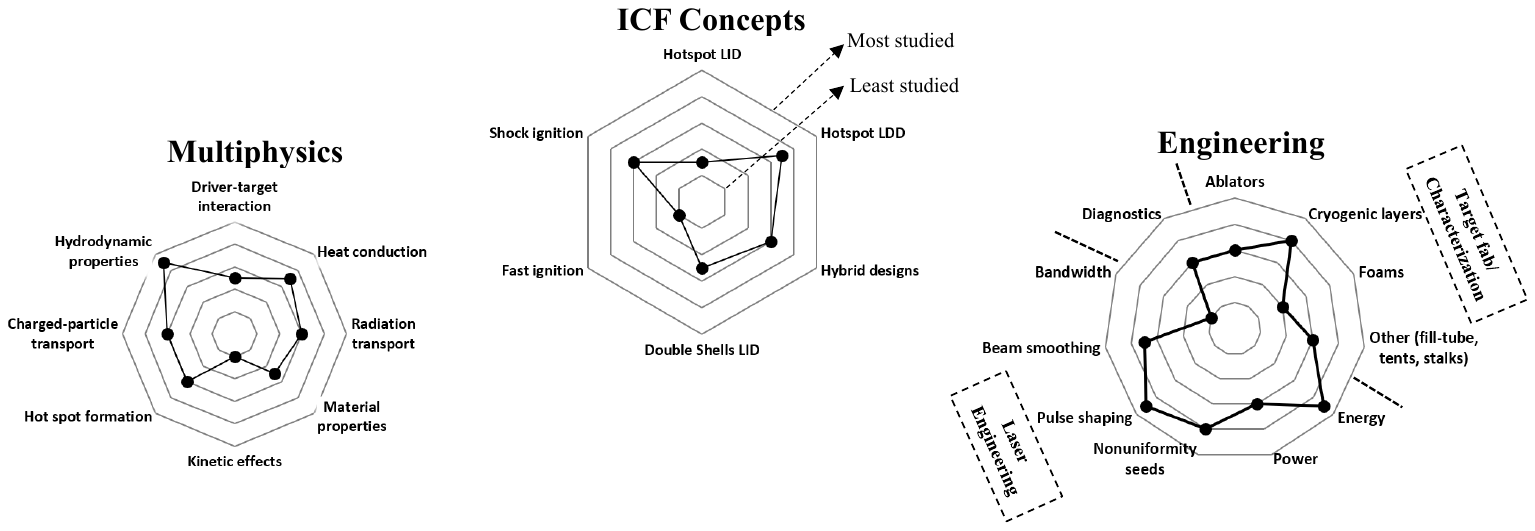}
\vspace{-100pt}
\caption{Notional chart indicating the three broad categories of subjects - concepts, the science associated with each concept, and the engineering aspects. Most studied refers to the largest set of experiments devoted to the particular concept or how well understood the concept is. Conversely, least studied are concepts with fewer studies experimentally or theoretically. 
\label{fig:SpiderPlot}
}
\end{figure*}

This description represents only one region of a much broader ICF design space. Fig.~\ref{fig:SpiderPlot} notionally illustrates various aspects of ICF. As the figure shows, ICF encompasses multiple concepts, including hotspot ignition \cite{Atzeni,Lindl,Craxton}, volume ignition \cite{LindlBook,AmendtDoubleShell}, fast ignition \cite{BasovFastIgnition,TabakFastIgnition}, and shock ignition \cite{ShockIgnition}. Evaluating any of these concepts requires predictive understanding of interacting phenomena such as laser--plasma interactions, hydrodynamic instabilities, radiation transport, thermal conduction, nuclear burn, and material properties including equations of state and opacities. Engineering constraints further couple these physical processes to laser delivery, capsule fabrication, target quality, and diagnostic capabilities. Experimental opportunities at facilities such as the National Ignition Facility (NIF) \cite{NIF} and OMEGA \cite{OMEGA} are necessarily limited in cadence, making ground-truth experimental data substantially scarcer than the data available in many scientific domains commonly used to develop and evaluate SI systems. In addition, multimodal data (scalar, vector and raster data) have uncertainties that are not always well quantified. Thus ICF presents an opportunity to test whether LLMs are capable of integrating knowledge across disparate physics regimes, with limited and uncertain data, and assisting experts. 

Despite these constraints, the field has made significant experimental progress, culminating in laboratory ignition \cite{Ignition} and target gain greater than unity \cite{Gain}. These achievements resulted from improvements in target design \cite{NIFTargetDesign}, diagnostics \cite{NIFDiagnostics}, laser performance \cite{LaserDelivery}, target fabrication and quality \cite{TargetQuality}, and the models used to interpret and design experiments. 

These characteristics make ICF a useful test case for examining pathways to improved language-model performance in specialized scientific domains. We compare off-the-shelf models, a model specialized through combined domain-adaptive pretraining (DAPT) and supervised fine-tuning (SFT) called Prospero, and retrieval-assisted models. Combined DAPT and SFT improves performance relative to the base model, but requires substantial computational resources and expert data curation; our study does not isolate the effect of DAPT versus SFT. Retrieval provides an alternative means of supplying domain information and improves performance for the smaller model tested. These findings motivate further investigation of models combining retrieval-augmented generation (RAG), scientific tools, and reinforcement learning (RL). In such systems, experimental observations—including failures and null results—could inform retrieved evidence and, where suitable objectives can be defined, verifiable rewards or feedback. We propose this combination as a future research direction.

This paper is organized as follows: A physics-based benchmark for evaluating LLMs using ICF and high-energy-density-physics ~\cite{HEDP} as the test domain is described in Section~ \ref{sec:ICFBenchmark}. The evaluation methodology is discussed in Section~\ref{sec:EvaluationMethodology}.  Comparisons of model endpoints including the base and the ICF-adapted models and performance analysis are presented in Section \ref{sec:Results}. Lessons learned and a potential path forward are discussed in Section~\ref{sec:Discussion}. Conclusions are presented in Section~\ref{sec:Conclusions}. Specific examples comparing Prospero with its base model are shown in Appendix A. Two detailed physics examples that serve as stress tests of the models are found in Appendices \ref{app:AppB} and \ref{app:AppC}. 

\section{\label{sec:ICFBenchmark} {ICF Benchmark}}
\begin{figure*}[ht!]
\includegraphics[trim=0cm 2.1cm 0cm 1.5cm, clip,width=0.9\textwidth]{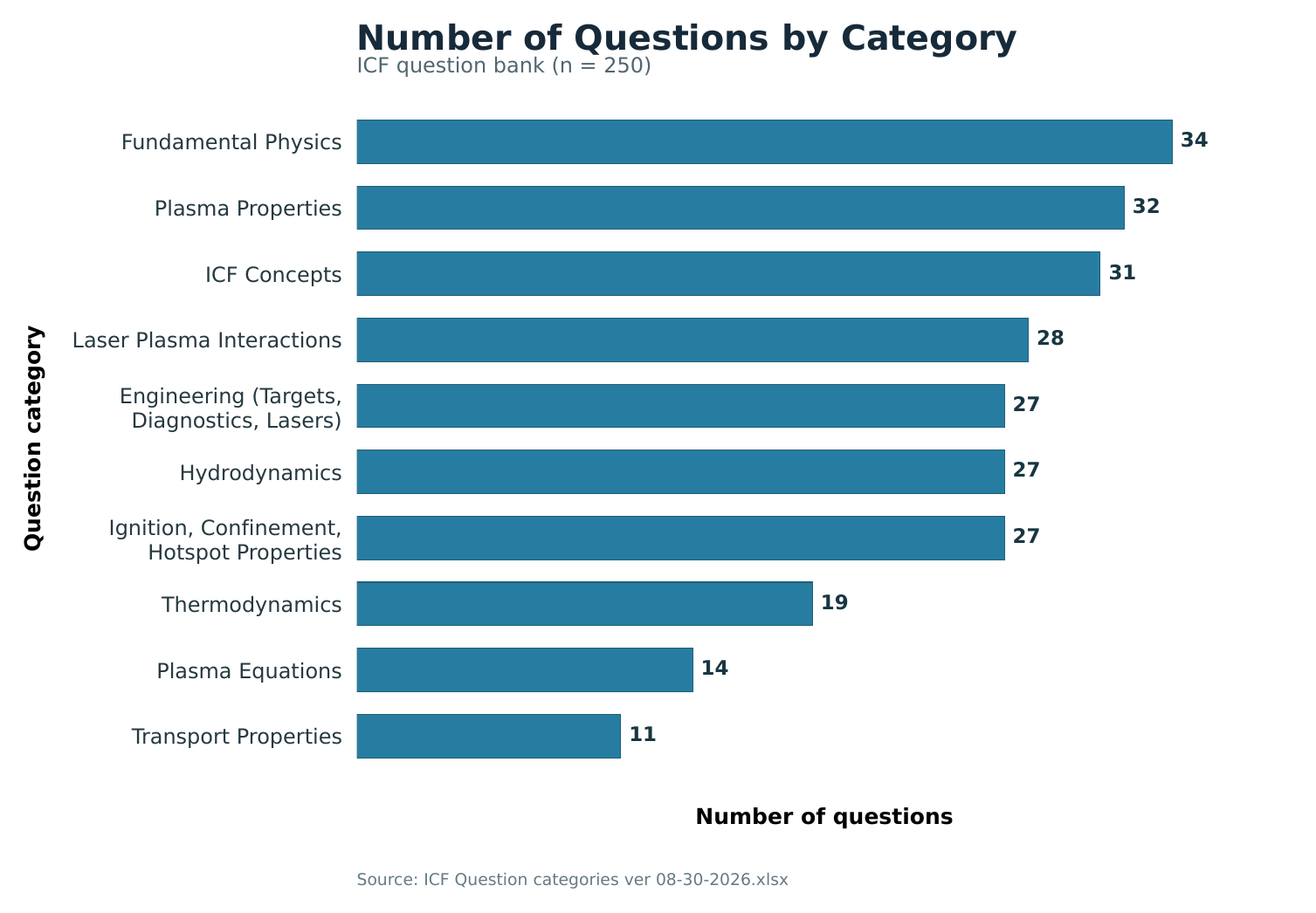}
\vspace{0pt}
\caption{Categories of questions developed for the golden-set - the set of 250 question and answer pairs generated by ICF experts. The numbers at the end of the bars indicate the number of questions of each type in the golden-set.
\label{fig:QuestionCategories}
}
\end{figure*}

With input from domain experts, we developed a ``golden-set" of 250 question-and-answer pairs. In principle, this set of benchmark questions can be used to assess any LLM intended for scientific applications. Questions in the golden-set cover a broad range of topics, from fundamental physics to questions requiring more specialized high-energy-density-physics knowledge including ICF. The distribution of questions across the different categories is shown in the histogram in Fig. \ref{fig:QuestionCategories}.

The benchmark questions are intentionally heterogeneous and require models to generate answers to physics problems across a range of topics and levels of specialization. Each question and its corresponding answer were curated by two independent ICF researchers and where their assessments conflicted, we resolved the disagreements through discussion by rephrasing or adding additional context. At the same time, we were careful not to provide so much context that the expected answer was effectively contained in the question itself. The goal was to create scientifically meaningful questions that required the model to draw on its awareness of the underlying physics and, where appropriate, reason across multiple physical processes.

Table \ref{tab:TopicsTable} lists examples of the topics covered within each category. Many of the physical processes that determine ICF implosion performance are coupled. Therefore, the categories are not mutually exclusive and the same topic can appear in several categories. For example, fuel compression is a key measure of implosion performance, but the extent of compression achieved depends on many aspects of the physics, including shock strength, equation-of-state and thermodynamic properties, laser-plasma interactions that influence the timing and sequence of shocks, hydrodynamic instabilities, and preheat from radiation or energetic electrons. As shown in Table \ref{tab:TopicsTable}, compression therefore appears as a relevant topic in several of the question categories. This overlap is representative of the multiphysics nature of ICF. 

\begin{table*}[ht]
\centering 
\renewcommand{\arraystretch}{2.0} 
\resizebox{\textwidth}{!}{%
\begin{tabular}{|c|c|} 
\hline
\textbf{Question categories} & \textbf{Example topics} \\ 
\hline
Fundamental physics & fusion reactivities, waves, rocket equation, electromagnetic forces \\ 
\hline
Plasma properties & plasma frequency, collision frequency, Debye screening \\ 
\hline
ICF concepts & hot spot ignition, fast ignition, volumetric ignition, aneutronic fusion \\ 
\hline
Laser plasma interactions & cross beam energy transfer, two plasmon decay, stimulated Raman scattering, preheat, compression \\ 
\hline
Engineering (targets, lasers, diagnostics) & instability seeds, characterizing laser speckle, laser optics, neutron diagnostics \\ 
\hline
Hydrodynamics & hydroefficiency, shocks, instabilities, compression, energy scaling \\ 
\hline
Ignition, Confinement, Hotspot properties & confinement time, hotspot dynamics, ignition criteria \\ 
\hline
Thermodynamics & shocks, Rankine-Hugoniot conditions, compression \\ 
\hline
Plasma equations & Boltzmann equation, diffusion equation \\ 
\hline
Transport properties & heat conductivity, kinetic effects, radiation transport, charged-particle transport \\ 
\hline
\end{tabular}
}
\caption{Examples of topics within each area of the questions in the golden-set. Note that some topics can be listed in multiple categories.}
\label{tab:TopicsTable}
\end{table*}

Questions fall into the types shown in Table \ref{tab:QuestionTypes}. We deliberately avoided multiple-choice questions as they measure answer selection rather than answer generation, though we do have one such question in the benchmark. Qualitative questions include those that require inferring hypotheses for certain experimental observations explicitly outlined in the question. These types of questions may also require drawing conclusions from equations or inferring trends from equations. Questions that require a numerical answer may be based on well-known equations that are unspecified in the question or answers to other questions in the set where a symbolic derivation is required. In the latter case, to be self-contained, the equation is specified in the numerical question. The symbolic derivation questions specify assumptions and require the model to follow a chain-of-thought during the derivation. Equation formulation merely asks to list the relevant equation, for example, the scaling of fusion reaction cross section with energy in the center-of-mass frame of two ions.

\begin{table}[ht]
\centering 
\renewcommand{\arraystretch}{2.0} 

\begin{tabular}{|c|c|} 
\hline

\textbf{Question Type} & \textbf{Percent in set} \\ 
\hline
Qualitative/Reasoning & 29.2 \\ 
\hline
Numerical calculations & 25.6 \\ 
\hline
Symbolic derivation/Reasoning & 24.0 \\ 
\hline
Equation formulation & 20.8 \\ 
\hline
Multiple-choice & 0.4 \\ 
\hline
\end{tabular}
\caption{Percentage of questions of each type within the golden-set of 250 questions}
\label{tab:QuestionTypes}
\end{table}

\section{\label{sec:EvaluationMethodology} Evaluation Methodology}

\begin{figure*}[ht!]
\includegraphics[trim=0cm 0cm 0cm 0cm, clip,width=1.0\textwidth]{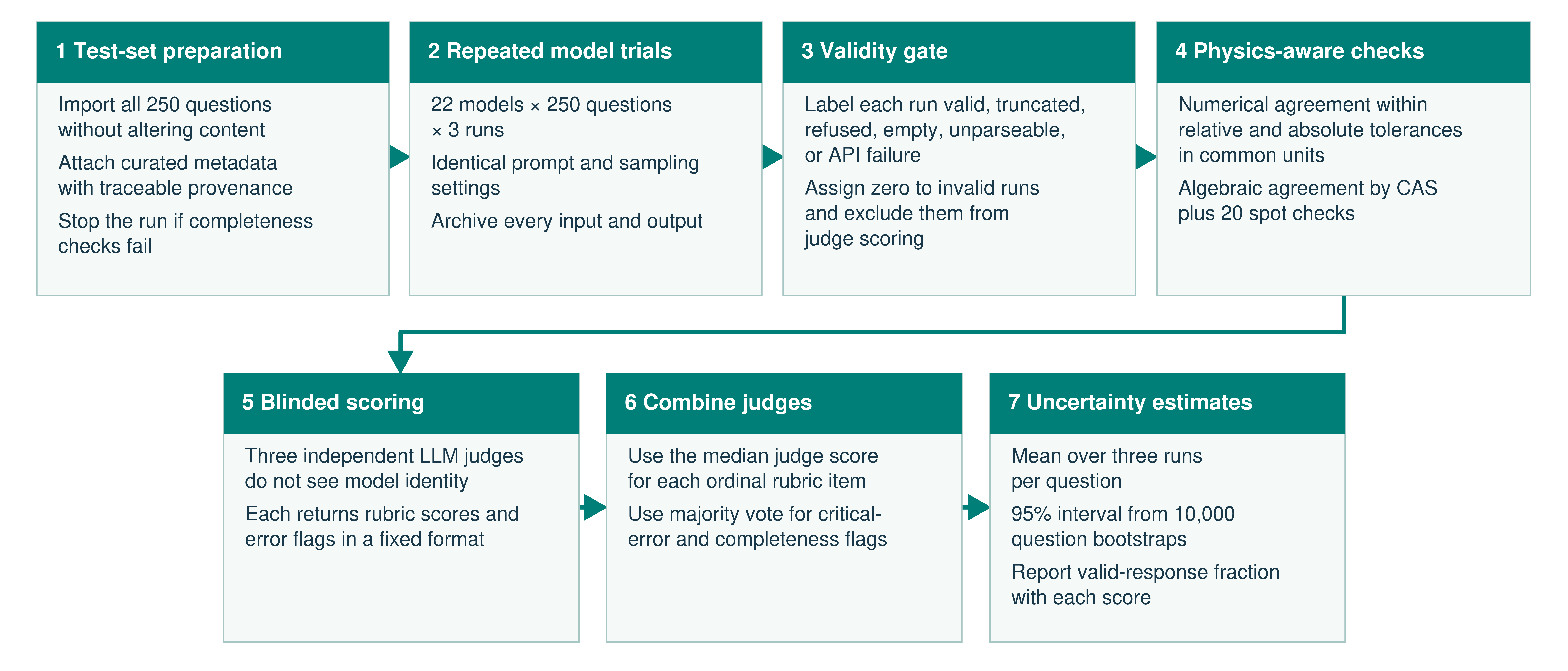}
\vspace{-10pt}
\caption{Workflow used to assess LLM outputs. Blocks 4 and 5 together form an effective single scoring stage: Block 4 lists the physics-aware equivalence criteria (numerical tolerance, unit consistency, algebraic equivalence) that the judges in Block 5 apply when comparing each answer with the answers in the golden-set. API: Application Programming Interface; CAS: Computer Algebra System; 10,000 bootstrap samples: statistical uncertainty estimated from 10,000 resamples of the fixed scores.
\label{fig:AIEvaluationWorkflow}
}
\end{figure*}

We established the pipeline shown in Fig.~\ref{fig:AIEvaluationWorkflow} to evaluate 22 models using the golden-set - each model is evaluated by three LLMs as judges on three independent attempts per question, for 16,500 planned answer calls. Since results from these models are probabilistic, we required more than one attempt to assess their performance; three was chosen as an ad hoc number. Models were evaluated without internet access or external literature retrieval. Examples of models are shown in Fig.~\ref{fig:leaderboard}. The evaluation ranges from off-the-shelf closed-weight models such as Claude Sonnet 5 \cite{Claude5}, Claude Sonnet 4.5 ~\cite{Claude4.5}, and GPT-5.4 ~\cite{GPT-5.4} to models with open weights including the NVIDIA Nemotron-Ultra-550B ~\cite{Nemotron-Ultra}, gpt-oss-120B ~\cite{gpt-oss-120B}, NVIDIA Nemotron Super ~\cite{NemotronSuper}, Inkling Small ~\cite{inkling}, Muse Glimmer ~\cite{MuseGlimmer}, and Gemma 4 \cite{gemma4}. The pipeline first imports the golden-set without altering content. It then performs the repeated trials of each question for each of the models. Each answer goes through a gate, where it is assessed for its validity [e.g. valid, truncated, refusal, empty, parse failure, or an error in calling the tools necessary to answer the question, i.e. an application programming interface (API) error].   Non-valid responses receive zero and are not sent for content judging. Then physics checks are performed - valid numerical answers are checked using relative and absolute tolerances, with units assessed separately and multiple values matched one-to-one. Scientific errors such as unit errors, using improper dimensions in equations, or incorrect equations are flagged as critical errors. Symbolic answers are tested by a computer-algebra system (CAS). Based on the rubric, each judge LLM assigns a score for each question. As mentioned earlier, a majority vote is used for responses flagged as critical errors. The scores are averaged over three attempts per question; all questions are weighted equally to calculate the mean.  Keeping the calculated scores fixed, we bootstrap the statistical uncertainty in the score, i.e. we generate 10,000 new samples by randomly drawing from these scores with replacement, allowing each score to be selected more than once. The variation in the result across these samples provides an estimate of its statistical uncertainty, expressed as a 95\% confidence interval.
The same process is applied to the ICF-adapted fine-tuned model and the retrieval study (Sec.~\ref{subsec:retrieval}). Results from this evaluation process are shown in Section~\ref{sec:Results}. 

\subsection{\label{subsec:LLMJudges}Blinded, physics-aware judging} 
Valid golden-set answers are evaluated by three independent judges: GPT-5.4, Mistral Medium 3.5~\cite{Mistral-Medium-3.5}, and NVIDIA-Nemotron-Ultra-550B. The assessments are blinded, i.e. the judges are unaware of the models that generated the answers being assessed. They separately assess factual correctness, causal explanation, relevance, completeness, contradictory derivations, and critical physics errors as shown in Table~\ref{tab:Scoring Rubric}. 

 A single text-similarity metric cannot grade all four answer forms. Numerical answers may differ in units, rounding, or scientific notation while remaining physically equivalent. Symbolic expressions can be algebraically identical while seeming different, for example, $ln(a/b)$ is the same as $-ln(b/a)$ and the judges should be able to parse such expressions correctly. Equations may be equivalent up to a nonzero constant multiple, whereas multiplying a bare expression by a constant changes its value. Qualitative answers have no canonical wording and must be judged by whether they include the required claims, preserve the physics, and explain the relevant physics relationships. Thus, exact match and semantic similarity are retained as diagnostics and neither is allowed to override a physics-aware check. We used the median for ordinal and dimensional scores to limit the effect of one anomalous judge. For Boolean critical-error flags, we required agreement from two of the three judges. A critical error caps the score at 40\%, even if a final number happens to match, while a correct value supported by an invalid or contradictory derivation is capped at 70\%. Non-valid responses, including refusals, truncated outputs, empty responses, and parse failures, receive zero. This structure follows the broader motivation for rubric-based, multi-judge evaluation \cite{zheng2023,liu2023,kim2024}. 

\begin{table*}[ht]
\centering 
\small
\renewcommand{\arraystretch}{2.0} 

\resizebox{0.8\textwidth}{!}{%
\begin{tabular}{|c|c|} 
\hline
\textbf{Question Type} & \textbf{Rubric Weighting} \\ 
\hline
    Numeric & 55\% value; 15\% units and dimensions; 20\% method; 10\% assumptions and uncertainty. \\
\hline
    Symbolic & 60\% expression; 15\% signs, variables, and units; 20\% derivation; 5\% assumptions. \\
\hline
    Conceptual & 60\% required claims; 25\% factual correctness; 10\% causal explanation; 5\% relevance. \\
\hline
    Multi-part & Weighted mean of explicit subpart scores; every subpart is required for completeness. \\
\hline
\end{tabular}
}
\caption{Scoring rubric across each question type}
\label{tab:Scoring Rubric}
\end{table*}

\begin{figure}[ht!]
\includegraphics[trim=0cm 1.3cm 1cm 1.65cm, clip,width=0.8\columnwidth]{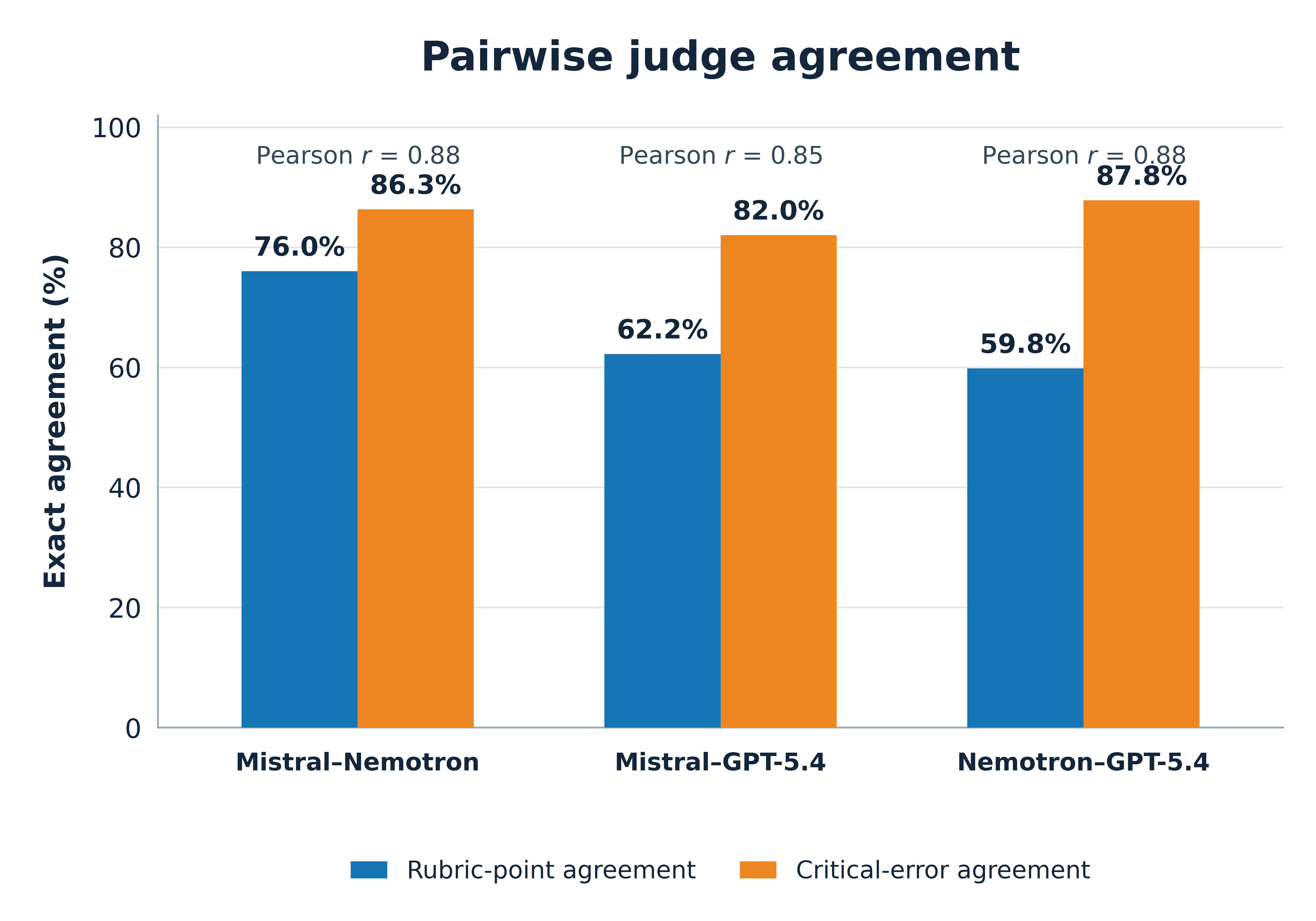}
\vspace{-10pt}
\caption{Pairwise agreement between judges on scores (blue) and on the critical-error rates (orange).
\label{fig:judge_agreement}
}
\end{figure}

Fig.~\ref{fig:judge_agreement} shows the pairwise agreement between judges. Pairwise rubric-point agreement ranges from 59.8\% to 76.0\%, while critical-error agreement is higher, from 82.0\% to 87.8\%. 
Because the judges rated the answers independently, the relatively high Pearson’s r coefficients for both the rubric and the critical-error flags suggest that they used the rubric effectively.

Using three judges reduces the influence of an anomalous score. For the example shown in Appendix B, Mistral as judge incorrectly awarded a score of 100 for one attempt, while the other two judges correctly assigned 40, as they found a critical physics error. Taking the response-level median prevented the isolated full-credit judgment from controlling the result. Multiple judges thus improve assessment robustness.

We assessed the judges on randomly chosen questions, one per category. Across ten categories and three judges assessing the same question, this resulted in a total of 30 judgements that we assessed. We agreed with the judges on all their assessments of these questions leading to further confidence in their capability, given the rubric. 

\begin{figure*}[ht!]
\includegraphics[trim=0cm 0cm 0cm 2.4cm, clip,width=0.8\textwidth]{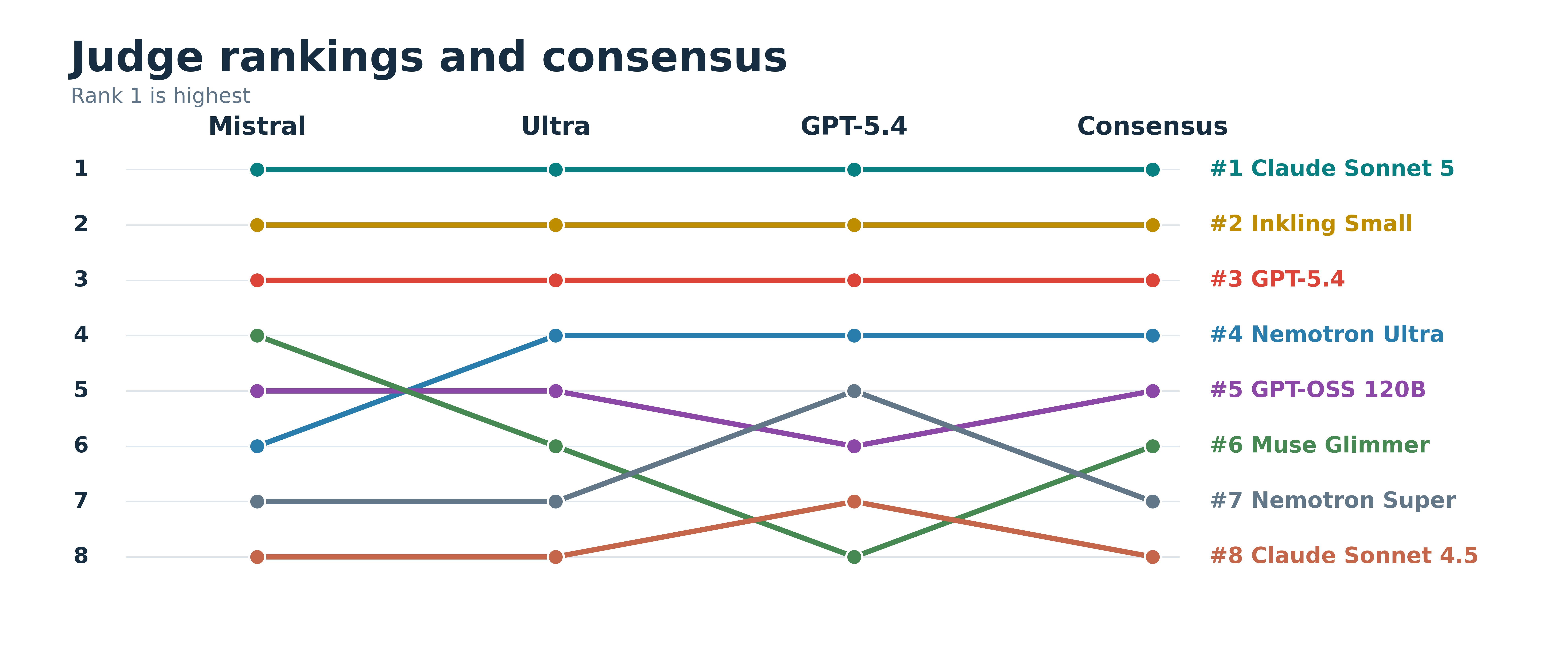}
\vspace{-10pt}
\caption{Stability of judge ranking. Consensus among the three judges is used for ranking. The top models are unanimously assessed as the best performing models. Models following those change places but never more than two spots. Only the top eight models of the 22 assessed are shown. Model references in text. 
\label{fig:judge_ranking}
}
\end{figure*}

The three judge families produce the same top-three model order, and no model in the consensus top ten differs by more than two positions of the consensus ranking among judges (Fig.~\ref{fig:judge_ranking}). In combination with the pairwise judge agreements, this consistency supports the broad ranking. 

\subsection{\label{subsec:DAPT} Domain-adaptive pretraining (DAPT) and supervised fine-tuning (SFT)}
\begin{figure*}[ht!]
\includegraphics[trim=0cm 0cm 0cm 0cm, clip,width=0.8\textwidth]{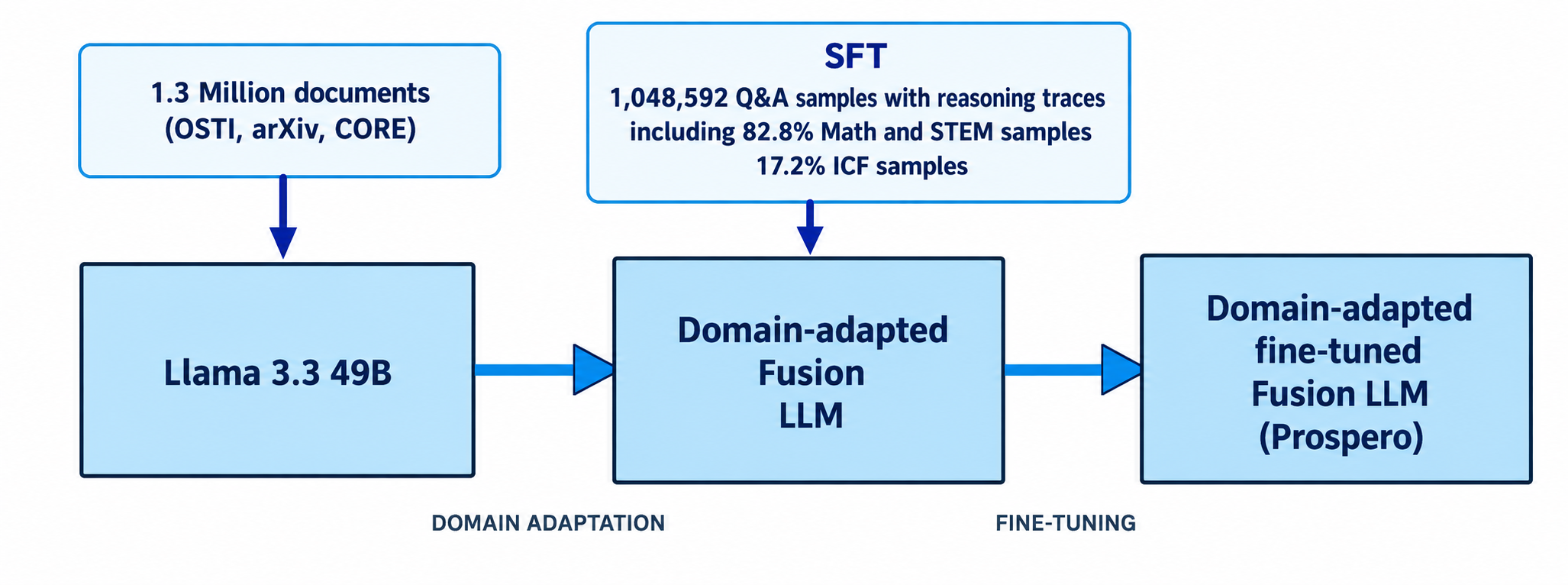}
\vspace{-10pt}
\caption{Process followed for Domain adaptive pretraining (DAPT) and supervised fine-tuning (SFT)
\label{fig:Prospero}
}
\end{figure*}

A separate specialization study summarized in Fig.~\ref{fig:Prospero} was conducted using the open-weight Llama-3.3-Nemotron-49B-Instruct-128K model \cite{nvidianemotron}. DAPT exposed it to approximately 1.3 million scientific documents, corresponding to about 17 billion tokens from arXiv.org \cite{arxivDatabase}, the U.S. Department of Energy Office of Scientific and Technical Information \cite{ostiDatabase}, and the CORE \cite{coreDatabase} databases. These papers were identified using keywords typical of ICF. Since many papers appear in multiple places, leading to erroneous weighting of documents, a de-duplication process was followed to ensure that each paper appeared only once in the corpus visible to the model. As part of the DAPT process, the model was systematically exposed to tokens (continuous pretraining or CPT) to study performance on standardized benchmarks such as GPQA \cite{GPQA}.  Subsequent SFT used 1,048,592 total samples, all with reasoning traces. Of these, 868,339 (82.8\%) were Math and STEM Q\&A pairs, while the remaining 180,253 samples were ICF Physics Q\&A pairs. Papers used to generate the ICF Q\&A pairs were not used during the DAPT process. Of the ICF samples used for SFT, a subset of 650 Q\&A pairs was curated by ICF scientists. Many questions and/or answers were rewritten. Some were eliminated as poor questions resulting in a net of 450 question-and-answer pairs that were deemed rigorous and pertinent. This ratio was used on the larger set of questions for SFT. A general chat instruction set was also applied to the model. Following these, a reasoning and instruction-following data blend was used to recover general capabilities. A longer discussion of this process is beyond the scope of this paper.

\begin{figure}[ht!]
\includegraphics[trim=0cm 3.5cm 0cm 4cm, clip,width=0.8\columnwidth]{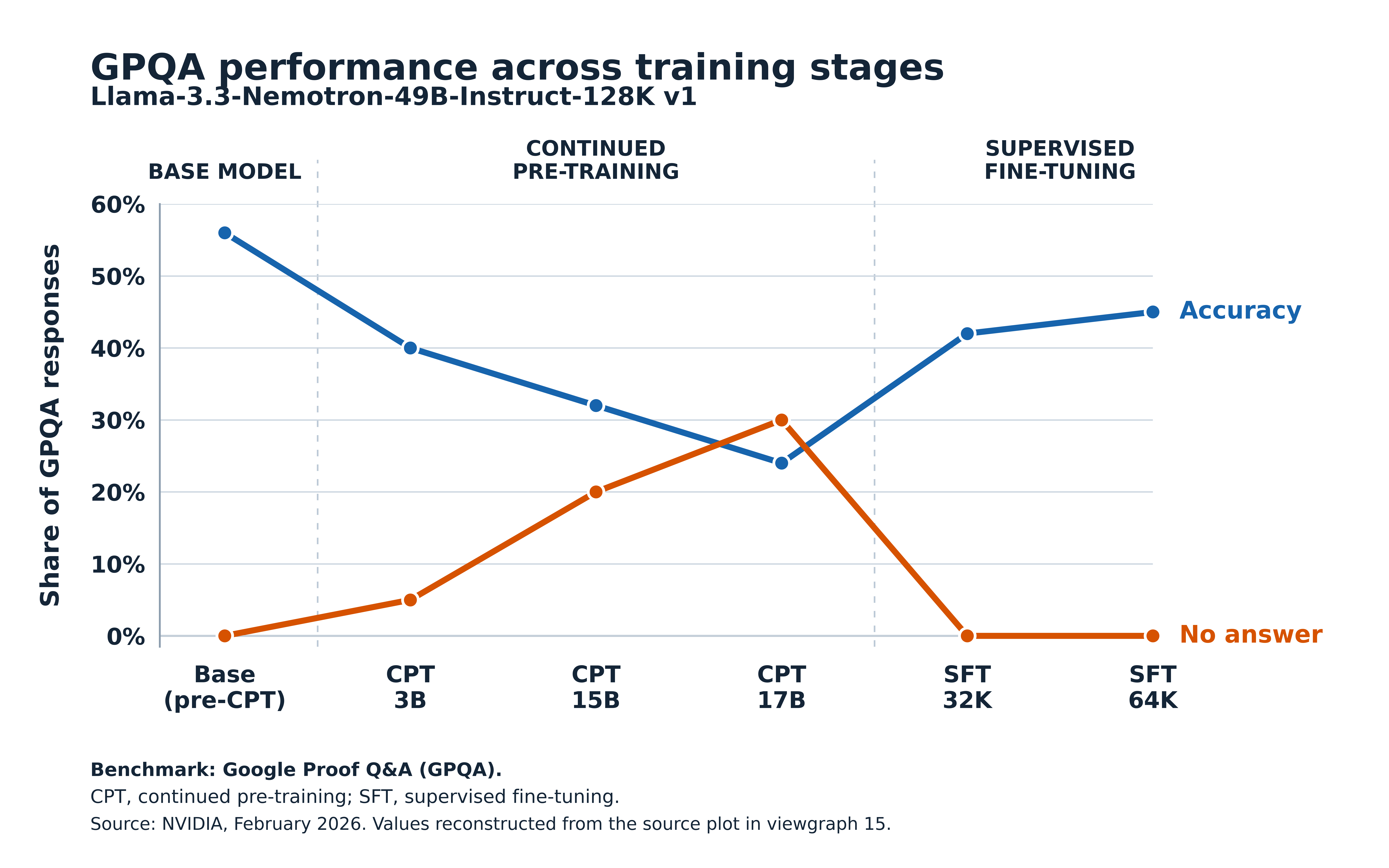}
\vspace{-10pt}
\caption{Example of an intermediate stage during the training process for the standardized benchmark GPQA. With increasing ICF training or continuous pretraining (CPT) performance drops (blue line) but can be recovered with supervised fine-tuning (SFT). The drop in performance is correlated with the increasing percentage of questions that are unanswered (orange line). Accuracy is recovered when the percentage of unanswered questions drops to zero.
\label{fig:GPQA}
}
\end{figure}

We tracked the cost of specialization during the CPT process with standardized benchmarks. Fig.~\ref{fig:GPQA} shows the accuracy in answering questions for the Graduate-Level-Google-Proof Q\&A benchmark (GPQA) (as an example) versus increasing number of ICF tokens, followed by a fine-tuning process. General-science performance drops with increasing number of ICF-related tokens. This correlates with the dominant reason for the loss of accuracy: increasing percentage of questions unanswered by the model with increasing CPT (orange curve). However, performance improves after broader SFT data are restored. Note that this figure is for an intermediate training strategy. An optimal blend of SFT data was subsequently identified that led to a nearly 16\% improvement in GPQA performance for the DAPT and SFT model, Prospero, relative to the base model for GPQA (not discussed further in this paper). Similar improvements were observed for other standardized benchmarks. We discuss Prospero's performance on the golden-set in the next section.

\section{\label{sec:Results} Results}

\subsection{\label{subsec:AggregatePerformance} Aggregate model performance}
\begin{figure*}[ht!]
\centering
\includegraphics[trim=0cm 0cm 0cm 3cm, clip,width=0.9\textwidth]{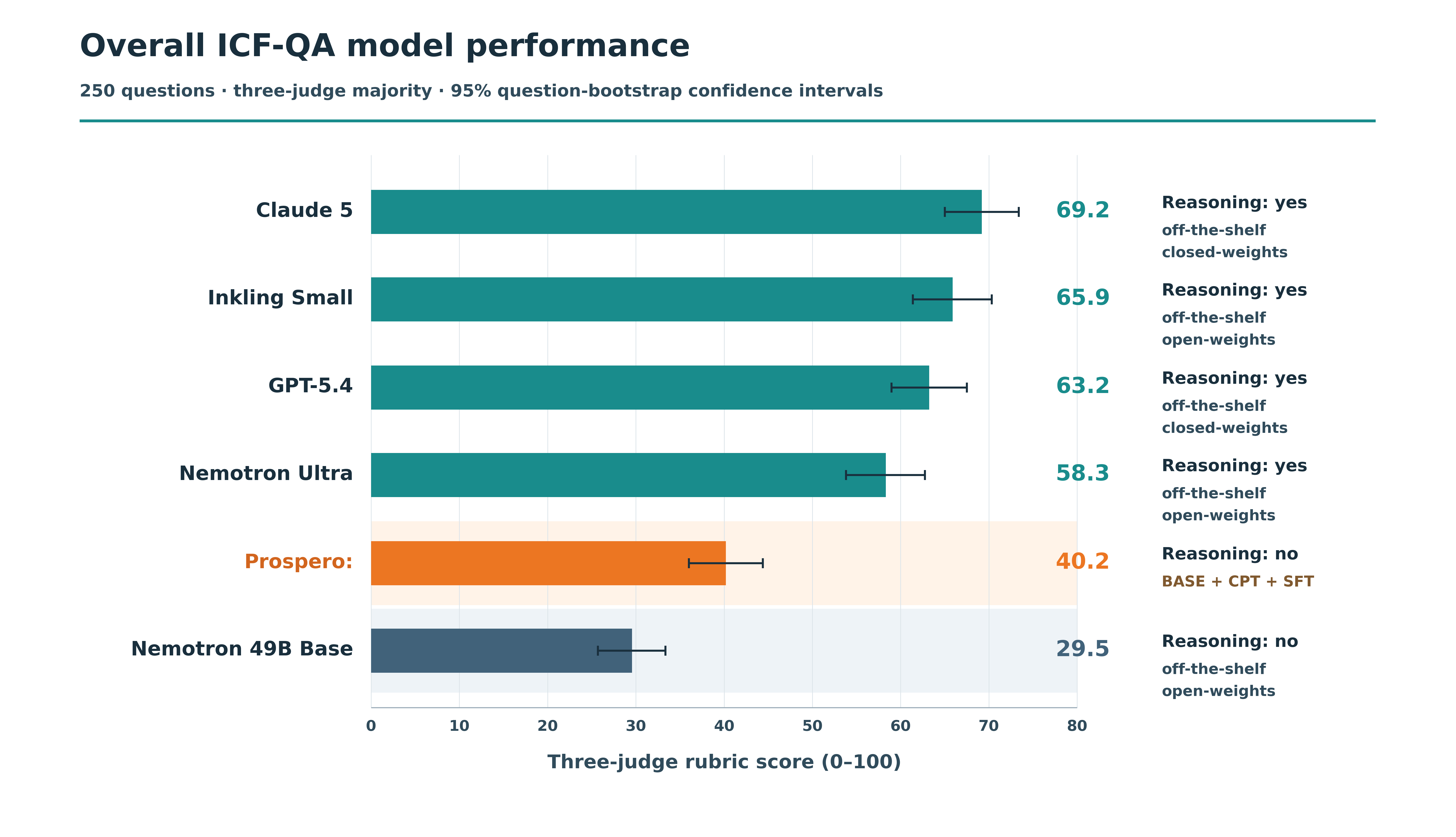}
\caption{Overall performance of the top 4 models, Prospero and its base (Llama-3.3-Nemotron-49B-Instruct-128K) used in this work on the 250-question ICF benchmark. Bars show the three-judge rubric score; horizontal intervals are 95\% confidence intervals from a 10,000-replicate question bootstrap. Each question contributes the mean of three attempts, and missing attempts count as zero.}
\label{fig:leaderboard}
\end{figure*}

The overall ranking in Fig.~\ref{fig:leaderboard} separates a small leading group from a long tail of substantially smaller/sub-frontier systems. The error bars in Fig.~\ref{fig:leaderboard} indicate the 95\% confidence interval. Claude Sonnet 5 has the highest coverage-adjusted rubric score, 69.2.  Inkling Small follows at 65.9 followed by GPT-5.4 at 63.2 and Nemotron Ultra with a score of 58.3 rounding out the top four. The next five models (not all shown) have overlapping intervals, so there is no stable ranking beyond the top four models. These results should not be interpreted as a leaderboard for the ICF application. Answers on these questions do not imply domain expertise. Moreover, rankings are ephemeral as models evolve. This ranking is interesting because of the improvement in the Prospero model. As the figure indicates, Prospero's performance (orange) improves 10.7 percentage points over the base model (grey) with a relative 36\% improvement over the base model. This indicates that DAPT + SFT could be a valuable strategy to improve the performance of small models on domain-specific tasks. We will discuss challenges and the relevance of DAPT and SFT in Section~\ref{sec:Discussion}. 

\subsection{\label{subsec:SegmentedPerformance} Segmented Model Performance}
\begin{figure*}[ht!]
\centering
\includegraphics[trim=0cm 6cm 0cm 3cm, clip,width=1.0\textwidth]{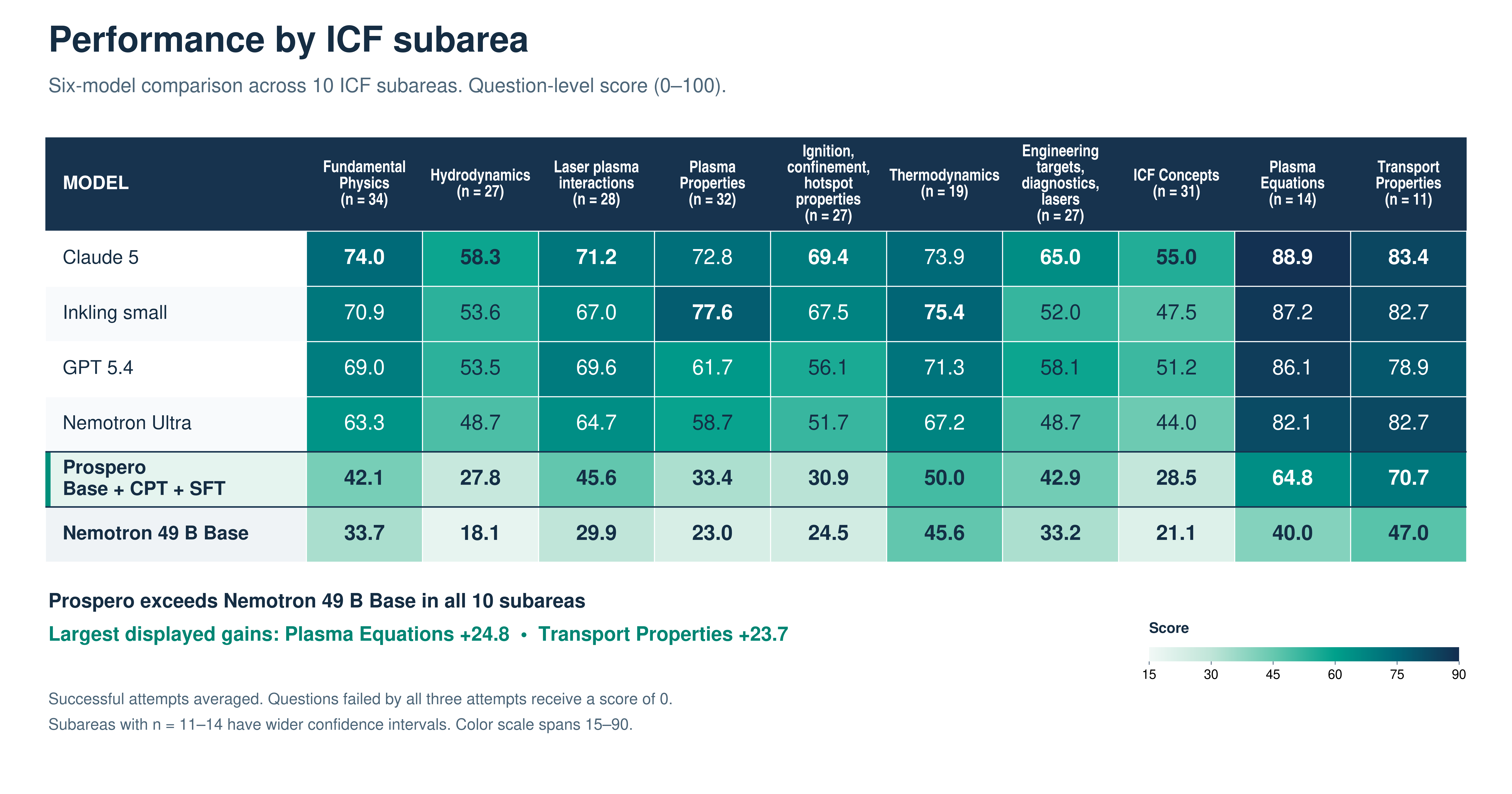}
\caption{Heat map showing scores of the top 4 models on the golden-set by ICF subarea/category compared with Prospero, and its base model (Llama-3.3-Nemotron-49B-Instruct-128K).}
\label{fig:PerformanceByArea}
\end{figure*}

When answers are analyzed by category, shown in Fig.~\ref{fig:PerformanceByArea},  Claude Sonnet 5 leads eight of the ten ICF subareas, while Inkling Small leads Plasma Properties (77.6) and Thermodynamics (75.4). The highest leader scores occur in Plasma Equations (88.9) and Transport Properties (83.4), but these are also the two smallest categories, with 14 and 11 questions respectively, and consequently have wider intervals. ICF Concepts (55.0), Hydrodynamics (58.3), and Engineering (65) remain the hardest categories even for the best system in each slice. Prospero exceeds the base in all categories with the largest gains in Plasma Equations (+24.8) and Transport Properties (+23.7).

\begin{figure*}[ht!]
\centering
\includegraphics[trim=0cm 5cm 0cm 3cm, clip,width=0.8\textwidth]{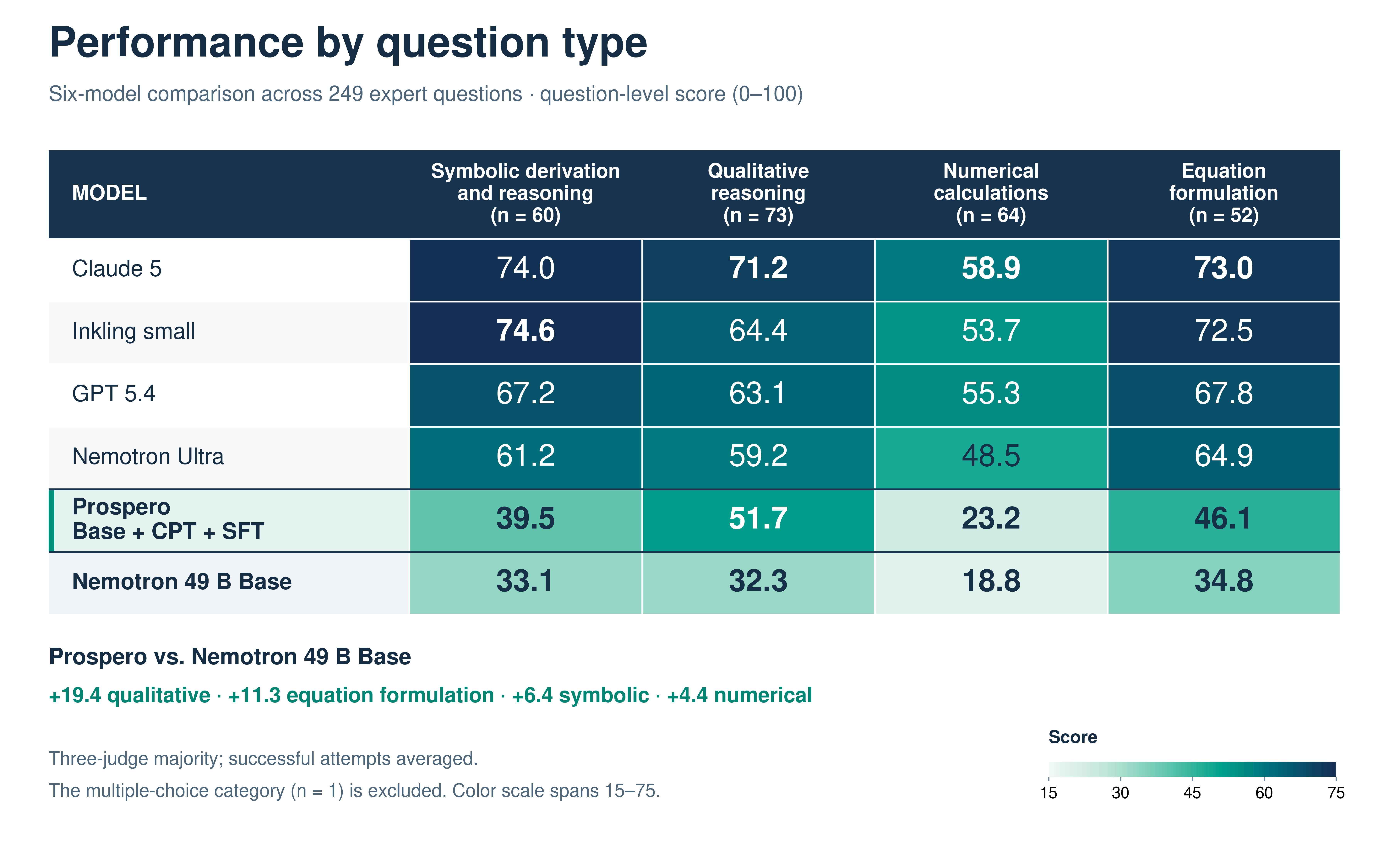}
\caption{Heat map showing scores of the top 4 models on the golden-set by question type compared with Prospero and its base model (Llama-3.3-Nemotron-49B-Instruct-128K). The single multiple-choice question is not included. }
\label{fig:PerformanceByType}
\end{figure*}

Performance changes when the same answers are grouped by question type (Fig.~\ref{fig:PerformanceByType}). Numerical calculations are the most difficult type for every model. Claude Sonnet 5 leads this slice at 58.9, but judges still identify critical errors in 58--71\% of numerical responses, depending on the answer model. Symbolic performance reveals a different ordering: Inkling Small leads at 74.6, while Muse Glimmer reaches 70.1 on symbolic derivation (not shown) but only 34.1 on qualitative reasoning. These contrasts show why a single benchmark average is inadequate for selecting a model for scientific work. A system that is strong at algebraic manipulation may still be weak at physical explanations, and the reverse can be true. To note, Prospero outperformed the base model on all question types: +18.4 on qualitative questions, +11.3 on equation formulation, +6.4 on symbolic reasoning questions, and +4.4 on numerical questions, indicating improvement across the spectrum of question types. 

\subsection{\label{subsec:ProsperoPerformance} Detailed performance of Prospero versus the base model}
\begin{figure*}[ht!]
\centering
\includegraphics[trim=0cm 0cm 0cm 1.6cm, clip,width=0.6\textwidth]{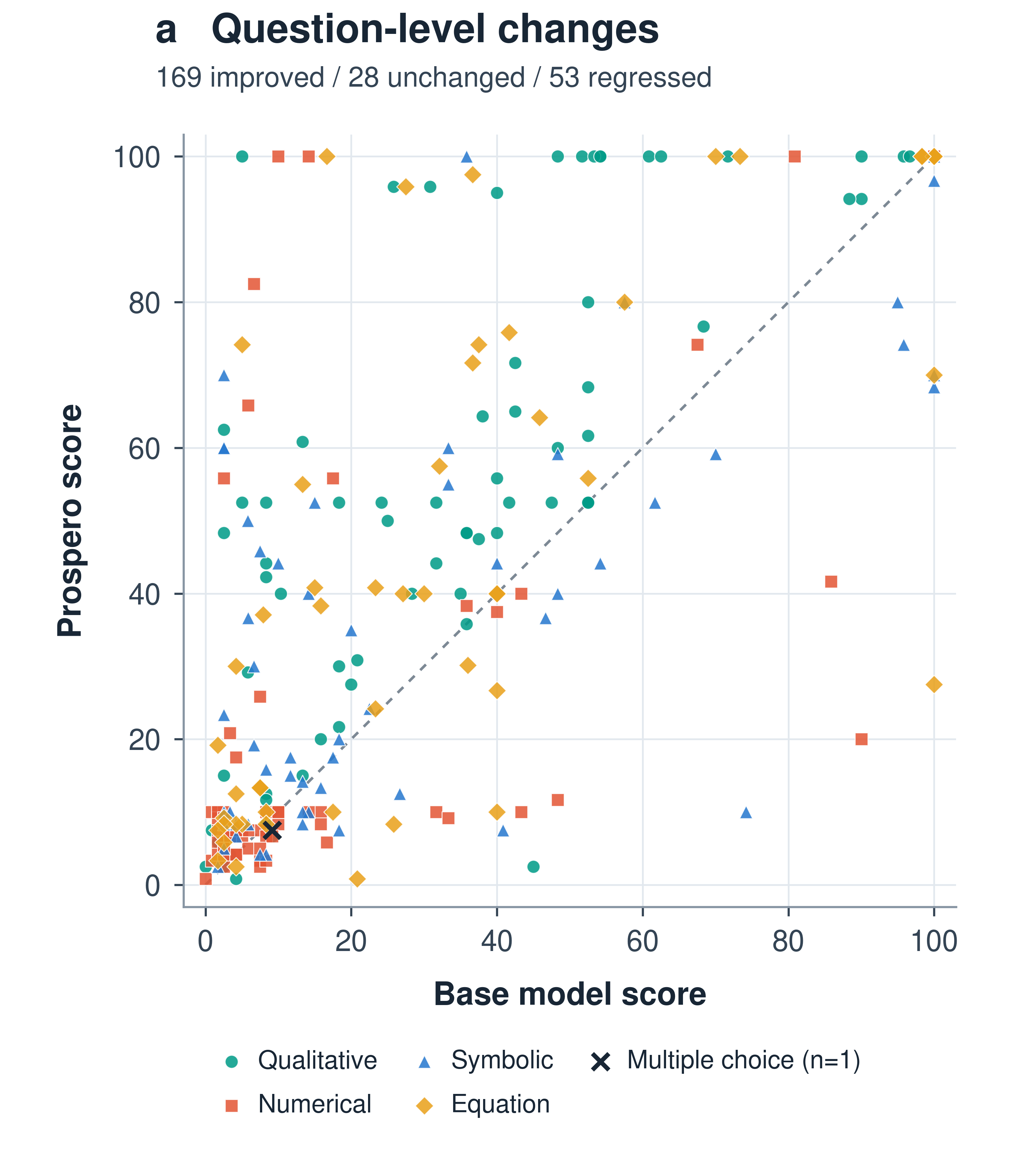}
\caption{Prospero performance on golden-set questions versus the base model. }
\label{fig:ProsperoQuestionLevelChanges}
\end{figure*}

In Sec.~\ref{subsec:AggregatePerformance} we described the aggregate performance of Prospero and identified 10.7 percentage points improvement over the base model. In Sec.~\ref{subsec:SegmentedPerformance} we showed that Prospero showed gains on all question types and categories. Here we decompose the results based on individual questions. Improvement is not uniform across the 250 questions (Fig.~\ref{fig:ProsperoQuestionLevelChanges}). Of the 250 questions, 169 questions (67.6\%) improved in their score, 28 questions remained unchanged (11.2\%), and 53 questions (21.2\%) regressed. Most of Prospero's improvements come from reduced physics errors. Examples of improved and regressed behavior are discussed in Appendix A. 

\subsection{\label{subsec:retrieval} Retrieval helps when it fills a knowledge gap}

\begin{figure*}[ht!]
\centering
\includegraphics[trim=0cm 0cm 0cm 0cm, clip,width=0.8\textwidth]{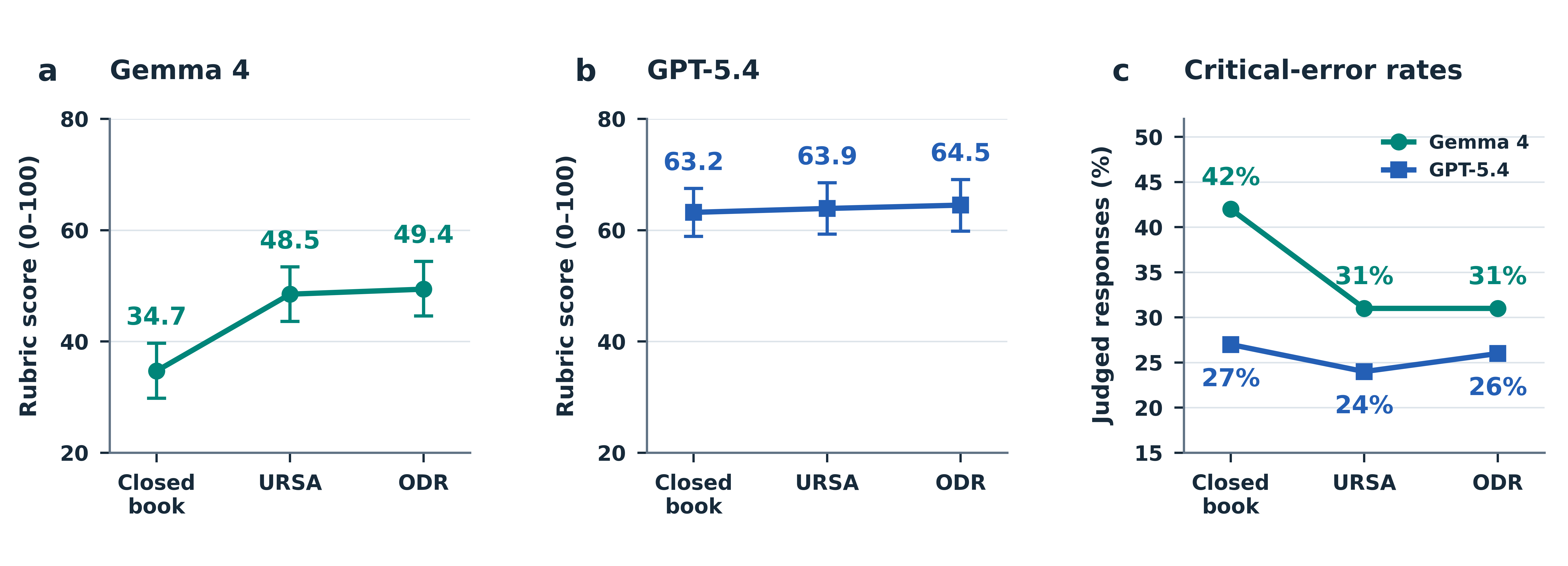}
\caption{Model performance for (a) the smaller Gemma 4 model; and (b) the larger GPT-5.4 model as a function of the literature available to the model. (c) Gemma 4 improves because its critical or physics errors drop with retrieval whereas GPT-5.4 performance does not improve significantly with additional retrieval provided by URSA or ODR. }
\label{fig:Retrieval}
\end{figure*}

 We tested whether poorer performance on the smaller models (not shown in the figures above) relative to the frontier models was due to their lack of domain knowledge.  We compared closed-book answering with tool-assisted answering on the same 250 questions and using the same judge panel. These results are shown in Fig.~\ref{fig:Retrieval}. For example, retrieval significantly improves the score of Gemma 4 \cite{gemma4}; its mean rises from 34.7 closed-book to 48.5 with the Universal Research and Scientific Agent (URSA) \cite{URSA} evidence and 49.4 with Open Deep Research \cite{ODR} evidence. The difference between the two retrieval systems is small, 0.94 points, with a paired 95\% interval spanning $-3.70$ to $+5.80$. Clearly, the data do not establish that either one is better. However, the improved performance, though less than the top four models in Fig.~\ref{fig:leaderboard}, indicates that smaller models benefit from retrieval. In this work we observed that Open Deep Research is about twice as fast per response in the comparison, but consumes approximately 3.4 times as many tokens. 

Retrieval provides much less benefit to the frontier GPT-5.4 model. Scores on the golden-set are 63.9\% with URSA evidence and 64.5\% with Open Deep Research evidence, compared with 63.2\% closed-book. Given the confidence intervals these variations are not statistically significant. 

\section{Discussion}\label{sec:Discussion}

\subsection{Limits of a domain benchmark} 
The golden-set metrics provide a useful measure of domain-specific performance but cover only a small fraction of the physical phenomena, regimes, and scientific questions relevant to ICF. Developing the benchmark required reducing a high-dimensional, multiphysics domain to tractable Q\&A pairs. In addition, expert-authored questions can introduce bias toward established knowledge, limiting representation of areas where the science remains uncertain. Consequently, some questions require qualitative rather than quantitative answers, making objective scoring more challenging.

Examples of the types of questions typically encountered by ICF scientists are shown in the Appendices. In Appendix~\ref{app:AppB} we identify a question that serves as a suitable stress test for the models. We additionally identify how to prompt LLMs towards the correct answer. In Appendix C, we discuss a challenging question that requires collating the relevant information from different references and performing calculations that span the multiphysics space of ICF. This example also illustrates that LLMs can be valuable scientific assistants, if they have access to the relevant tools including those for visual inspection, Python scripting, and symbolic manipulation. We conclude based on these and other representative examples that using LLMs effectively for science requires subject matter expertise.

\subsection{Benefits and costs of domain adaptation and fine-tuning}
DAPT and SFT come with several practical limitations.  The process of using peer-reviewed literature introduces publication bias toward successful or positive results and underrepresents null results and experimental failures. This is particularly important because these underrepresented experiments may contain some of the most valuable information for identifying knowledge gaps and enabling scientific discovery.

This work does not study DAPT and SFT separately. Thus, DAPT's value remains unknown without an SFT-only comparison. Performance declined during the intermediate training run but supervised fine-tuning recovered performance. This raises the question of whether DAPT is critical to the process and suggests that the additional compute and training complexity related to DAPT may not be justified when a model is already domain-aware.

 Additional challenges associated with the DAPT and SFT process included provenance and continual learning. Once information is incorporated into the model through DAPT, it becomes difficult to trace a particular response or scientific claim back to the source that influenced it. Continual learning is particularly challenging in an experimental field like ICF. Experiments are expensive, which means that new ground-truth data are generated relatively slowly. At the same time, experimental regimes continue to evolve, and lessons learned from earlier data may not hold under new conditions. New experiments can challenge or contradict our existing understanding. As a result, the problem is not simply that of adding new information, but of incorporating new experimental results while preserving what remains scientifically valid. This work illustrates the significant computational intensity required to adapt a model while retaining its basic scientific capability. 

The challenges of keeping a model up to date, however do not preclude its usefulness in pursuing new knowledge. Although predicting an appropriate ICF target design may require the latest experimental results, many tasks draw on knowledge developed over decades. A model with a broad understanding of the ICF literature could help researchers interpret experimental results, identify physical effects that might explain observations, and propose experiments or calculations to test those explanations. It could also help develop improved physics models for ICF codes by drawing on numerical methods and ideas from other fields. These contributions could be valuable even if the model's knowledge is not current. 

\subsection{\label{subsec:RAG} Retrieval and/or reinforcement learning versus weight adaptation}

Language models will be most useful in science if they can work in a coordinated fashion to help researchers with the scientific process - hypothesize, down select hypotheses, design and execute experiments, and perform post-shot analysis to further identify hypotheses. The key question is whether, given the available evidence, a model or combination of models can guide the selection of an experiment that yields new information or improves the experimental result. 

These limitations favor systems that combine language models with retrieved evidence, scientific codes, expert review, and ultimately experimental validation. This is particularly important in ICF, where the ground-truth is embedded in experimental data. This poses challenges: experiments are expensive, measurements are incomplete, and interpretation is often model dependent. In addition,  data are sparse and are often stored in forms that require extensive processing before they can be used for training. Experimental records typically lack consistent metadata. Subject-matter experts must remain responsible for approving physical claims and experimental decisions but this introduces human bias in the process. 

Retrieval-augmented generation (RAG) supplies the model with relevant information at inference time. For ICF, retrieval requires a multimodal corpus - journal articles, shot reports, target specifications, diagnostic calibrations, and simulation results. Retrieval from these sources can preserve provenance better than weight adaptation.

The Gemma 4 and GPT-5.4 comparison in Section~\ref{subsec:retrieval} suggests that RAG is particularly useful for smaller models or models with limited domain knowledge. It may be less important when a strong model already contains the relevant information, although retrieval permits access to recent results and local data. Thus, RAG may still be important if a model is domain-unaware or does not have the complete domain knowledge. 

RAG does not resolve inconsistencies in the evidence. Published ICF literature favors successful experiments, while null results and failed shots may be poorly documented or unavailable. Published literature may contradict previous work or include incorrect results. A physicist must therefore judge whether the retrieved evidence is relevant and mutually consistent. 

We did not evaluate reinforcement learning in this study. A useful next test of these models is to perform a study on the golden-set after RL on Prospero. Our ultimate goal is to develop a multi-agentic pipeline that can retrieve ICF shot records, run a prescribed physics calculation, and revise a target recommendation against a well-defined goal, such as an experimental design. An advantage of ICF is that predictions from such models can be checked on existing facilities.

Finally, we discuss the value of domain-adapted and fine-tuned open-weight models over closed models. For the ICF domain, of which many models are aware, the quality of the data, the fidelity of the computational tools, and the judgment of the physicists using the system are just as important as the choice of the model (including the tools that can be accessed by the model). In this work, token limits on closed models restricted their ability to handle complex, multi-step tasks related to the problem presented in Appendix~\ref{app:AppC}. Open-weight models offer greater control over context length, inference configuration, and infrastructure. It is, of course, very likely that DAPT and SFT will play an important role for proprietary domains. We suggest that researchers perform domain-relevant benchmarks to assess model capabilities before embarking on a computationally expensive DAPT and SFT process. 

\section{\label{sec:Conclusions} Conclusions}
Using inertial confinement fusion (ICF) as a test case, we developed an expert-curated, 250-question benchmark and a physics-aware evaluation procedure to assess language models across qualitative reasoning, numerical calculation, symbolic derivation, and equation formulation. This evaluation revealed differences in capability and failure modes that aggregate scores alone can obscure. Prospero, specialized through combined domain-adaptive pretraining and supervised fine-tuning, improved its score by 10.7 percentage points over its base model, with gains across all question types. These results demonstrate the benefit of the combined training procedure, although the contributions of its individual stages remain unresolved.

Improved benchmark performance, however, did not establish scientific reliability. Frontier models achieved the highest aggregate scores, yet critical physics errors persisted, particularly in numerical calculations. The scientific stress tests further illustrated the importance of recognizing physical regimes, identifying missing assumptions, and assessing model validity. Retrieval improved performance for the smaller model tested but provided no statistically significant improvement for the tested frontier model, indicating that its benefits depend on the model and task.

Together, these findings support evaluating existing domain capabilities, failure modes, and retrieval benefits before committing to substantial additional training. They also show that useful scientific assistance can coexist with consequential reasoning errors, making expert assessment essential. Future work will investigate whether retrieval, scientific tools, and reinforcement learning grounded in verifiable feedback can reduce these errors and improve performance on independent scientific tasks. 

\section*{Acknowledgments}
The authors acknowledge the contribution of a large group of scientists at the Los Alamos National Laboratory who provided questions for the golden-set. They also acknowledge Nathan Debardeleben and Dave Rich for providing valuable compute resources. SI was used in the following ways: to generate figures since most figures were the results of LLM analyses, to illustrate schematics, and to strengthen the abstract and conclusions. All text was proofread by the authors and figures examined for accuracy. The benchmark questions used in this study are available from the corresponding author upon reasonable request. This paper corresponds to LANL document\# LA-UR-26-29532. This work was supported by the U.S. Department of Energy through the Los Alamos National Laboratory. Los Alamos National Laboratory is operated by Triad National Security, LLC, for the National Nuclear Security Administration of U.S. Department of Energy (Contract No. 89233218CNA000001).

\bibliography{PRXreferences}

\appendix
\noindent\section{\label{app:AppA} Prospero performance comparison with base model}
Here we compare answers from Prospero and the base model answers for questions randomly selected for cases where Prospero showed improvement or regressed in a majority across all three attempts at answering each question. 
\noindent\subsection{Examples of improved performance of Prospero compared to base}
 The examples shown below have been randomly selected from the 67.6\% of questions that had improved answers, where the base gave the incorrect answer in a majority of three attempts and Prospero succeeded uniformly. They indicate that Prospero's domain awareness and reasoning improved compared to the base model.  \\
Question: \textit{ Two protons are 10 microns apart. No other particles or fields are present. How much stronger is the electrical force between the protons than the gravitational force?} \\
The base response introduced power-of-ten errors in both force calculations and reported a ratio eleven orders of magnitude too large. Prospero canceled the common inverse-square dependence and recovered the correct distance-independent ratio. Prospero identified that symbolic manipulation could be used and then performed the numerical calculation correctly.\\
Question: \textit{Assuming the plasmas have the same temperature \(T_i\), how many times higher is \(E_0\) for D-He\(^3\) fusion than for D-T fusion? \(E_0\) is the energy where the fusion reactivity is maximized and is given by \(E_0 = 4^{-1/3}{(E_GT_i^2)}^{1/3}\) and where \(E_G\) is the Gamow energy \(E_G = 2\pi^2 Z_1^2 Z_2^2{(e^2/\hbar c)}^2\mu c^2\), where \(\mu = m_1m_2/(m_1 + m_2)\) is the reduced mass.} \\
The base response approximated the nuclear masses inconsistently and obtained unequal reduced masses for D-He\(^3\) versus D-T. Prospero recognized that tritium and $^3$He have approximately equal reduced masses, so only the charge factor changes and the result is the cube root of four. In this example also, Prospero performed a symbolic manipulation and performed a numerical calculation correctly. 

\noindent Question: \textit{List the most important three-wave parametric instabilities that can occur in a non-magnetized plasma.}\\
The base model replaced two-plasmon decay with four-wave mixing, even while acknowledging that four-wave mixing is not intrinsically a three-wave process. Prospero supplied Stimulated Raman Scattering, Stimulated Brillouin Scattering, and Two Plasmon Decay as the answers and described the daughter waves for each instability. This answer demonstrates domain awareness by Prospero. 

\noindent Question: \textit{A target design for an inertial confinement system is defined in terms of the target radius R, shell thickness \(\Delta\), laser energy on target \(E\) and the length of the laser pulse, \(t\).
If \(R_1\) is the radius of a target with thickness \(\Delta_1\) on a facility that delivers a laser pulse of energy \(E_1\) and pulse-length \(t_1\), what are the values of the radius \(R_2\), shell thickness \(\Delta_2\), and pulse-length \(t_2\) on a facility that delivers laser energy \(E_2\)?} \\
The base model treated the problem as underdetermined and proposed a square-root radius scaling under an unsupported fixed-pulse assumption. Prospero applied hydrodynamic similarity: mass and energy scale with the cube of the linear dimension, while radius, shell thickness, and pulse duration all scale with the cube root of the energy ratio. This answer demonstrates domain awareness by Prospero. 

\noindent\subsection{Examples of Prospero regression compared to base}
 The examples shown below have been randomly selected from the 21.2\% of questions that had poor answers, where the base gave the correct answer in at least two attempts and Prospero regressed in at least two attempts. Prospero introduced physics or units errors in its answers. These indicate that improvements in domain awareness and reasoning are uneven across the golden-set.  
 
\noindent Question: \textit{Consider a gas of particles all moving with the same magnitude of velocity, \(v\), with the directions distributed isotropically. Assume the gas particles are collisionless and have a density of N particles per unit volume, with mass m. What is the collision rate per unit area of particles striking a wall?}\\
 The base model gave the standard kinetic-theory result, $N v/4$, in two attempts. Prospero was unstable: the first attempt moved through 1/2, 1/$\pi$, and $\pi$/2 before settling on 1/4, so the judges flagged the contradictory derivation; the second attempt answered $N v/\pi$ and the third attempt resulted in an answer of $N v/2$. Prospero's regression in this case indicates loss of reasoning capability or its incapability to retrieve the relevant information. 
 
\noindent Question: \textit{A spherical shell is accelerated by ablation on the outside of the shell. Find an expression for the shell velocity u, in terms of exhaust velocity \(u_{ex}\), and the mass function \(M(t)\) and \(M_0\) the initial mass \(M(t=0)\).}\\
The base model applied momentum conservation correctly and obtained $u = u_{ex} ln(M_0/M)$ in two attempts. Prospero double-counted the momentum carried by the ablated mass, producing $ u = (u_{ex}/2)(M_0^2/M^2 - 1)$ in the first attempt and $u = u_{ex}(M_0/M - 1)$ for the second attempt, so the logarithmic dependence was lost in every attempt, indicating loss of domain awareness.

\noindent Question: \textit{Assume a fully-ionized, ideal-gas equation-of-state with plasma density, \(\rho = AP/T\), where \(P\) is the plasma pressure, \(T\) is the plasma temperature, and \(A = m_{DT}/(1+Z)\) is the average-DT mass in g with \(Z=1\) fully-ionized. Assume that energy is deposited into the shell from some form of preheat and this deposited energy raises the temperature of the shell to \(T+\Delta T\). Use the expression for the degradation in the areal density as \(\frac{\rho R_{exp}}{\rho R_{no preheat}} = \frac{1}{1+\Delta T/T}\) and the increase in temperature as \(\Delta T = \frac{2A H_s}{3M_s}\), and canonical scaling laws for the pressure as \(P(MBar) = 2.2 \alpha [\rho(g/cc)]^{5/3}\) and \(P(MBar) = 120 (I_{15}^{2/3})\), where \(H_s\) is the shell internal energy in  joules, \(M_s\) is the shell mass, \(\alpha \sim \rho R^{-0.6}\), \(P(MBar)\) is the pressure in MBar, and \(I_{15}\) is the laser intensity in units of \(1 \times 10^{15}\) W/cm\(^{2}\). Areal density degradation can be written as
\[
\frac{\rho R_{exp}}{\rho R _{no preheat}} \approx \frac{1}{[1+0.012H_s/(\xi_s\alpha^{3/5}I_{15}^{4/15})]}
\]
How much energy, \(H_s\) is needed to degrade areal density by 40\(\%\), i.e.~the ratio of the measured areal density to the no-preheated value is 0.6? Assume \(\xi_s = 0.5\), \(I_{15} = 1\), and \(\alpha = 2\).}\\
 The base model reported the answer as 42 J, approximately the correct answer. However, Prospero left the unit unspecified and suggested MJ and the second attempt made a critical numerical error. Since all units were specified in the question, the answer unit should have been clearly definable. This is an example of a regression, where the arithmetic was applied nonuniformly and the reasoning was inaccurate. 

\noindent\section{\label{app:AppB} Identification gap between current state of LLMs and ICF Physics using Rayleigh-Taylor instability as an example} 
The Rayleigh–Taylor (RT) \cite{Rayleigh, Taylor, Chandra} instability is a well-known mechanism that can degrade implosion performance, making it a representative benchmark or an appropriate stress test for ICF modeling. Developing a credible RT model requires identifying the relevant physical regimes, deriving the relevant equations, and applying appropriate asymptotic reasoning.
We demonstrate that language models recognize the relevant physical concepts but fail to identify transitions between physical regimes, for instance ignoring nonlinearity.  However, reasoning performance improves after targeted instruction while maintaining broad scientific awareness. Language models are also capable of reverse-engineering a model given an answer. 

\noindent{Stress test prompt:} \\
\indent\textit{ A light fluid is supporting a heavy fluid against an acceleration, $g= 8.2 \times 10^{15} cm/s^2$ in the x-direction. The Atwood number is A = 0.5. There is an initial sinusoidal perturbation in the y-direction, $\delta x= \eta_0 sin(ky)$. The amplitude, $\eta_0 = 1 \ \mu m$ and the wavelength $\lambda  = 100 \ \mu m$. Note that $k= 2\pi/\lambda$. Write the necessary expressions for the amplitude $\eta$ as a function of time t. Then calculate the numerical value of the amplitude $\eta$ of the perturbation at two times, 1ns and 3.3ns.} \\

 The appropriate answer must a) motivate the initial-value problem, b) evaluate the dimensional scales without unit error, c) calculate the amplitude in the linear phase, d) recognize when the growth rate has become nonlinear, e) identify the dimensionality of the problem, and f) calculate the value of the late time amplitude. Thus, this case study separates formula recall from model selection. 
\subsection{Analytic solution}

For an inviscid, incompressible, single Fourier mode of wavenumber $k$, the linearized interface displacement $\eta$ satisfies
\begin{equation}
\ddot{\eta}=A g k\eta, \qquad
k=\frac{2\pi}{\lambda}, \qquad
\gamma=(A g k)^{1/2}.
\label{eq:rt_linear}
\end{equation}
The general solution is
\begin{equation}
\eta(t)=C_{+}e^{\gamma t}+C_{-}e^{-\gamma t}
\label{eq:rt_general}
\end{equation}
where the variables are defined in the prompt above and $C_+$ and $C_-$ are constants that depend on the initial conditions. 
The implicit assumption in this problem (based on the scientist's solution or ``frozen" solution) is that the interface is initially at rest, $\dot{\eta}(0)=0$. 
Thus 
\begin{equation}
\eta_{\mathrm{L}}(t)=\eta_0\cosh(\gamma t)
\label{eq:rt_cosh}
\end{equation}
where the subscript $L$ defines the linear amplitude. 

Note that the prompt was sufficiently vague that if a model motivated the exponential solution, that would also be acceptable. But for the rest of the solution, we will stay with the hyperbolic choice of the single mode.
The nonlinear amplitude can be reproduced in at least two different ways shown below. Both assume a transition time $t_s$, at which the linear solution transitions to a nonlinear phase where the amplitude grows at a constant velocity $v_{NL}$. 

The resulting amplitude is given by
\begin{equation} 
\eta(t) = 
\begin{cases} 
  \eta_0 \cosh(\gamma t)  & t \le t_s \\ 
  \eta_s + v_{\mathrm{NL}}(t - t_s)   & t > t_s 
\end{cases} 
\label{eq:rt_piecewise} 
\end{equation}

\begin{figure}[t]
\centering
\includegraphics[width=0.6\columnwidth]{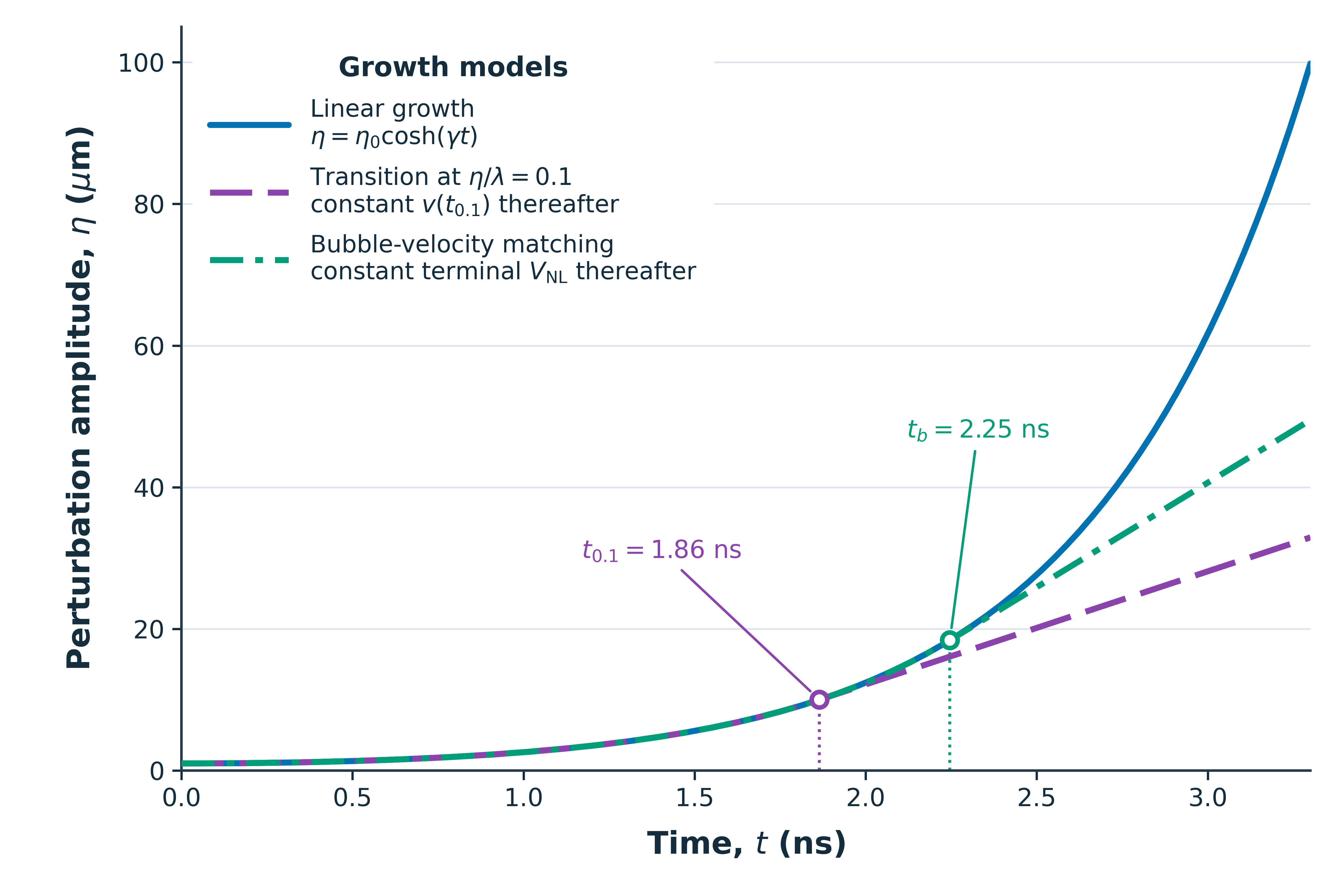}
\caption{Rayleigh-Taylor mode under three explicit assumptions. Purely growing hyperbolic cosine function (solid blue); amplitude from the piecewise method using the terminal bubble velocity $v(t_{0.1})$ given by Eq.~\ref{eq:haan} (dashed purple); amplitude from the piecewise method using the terminal bubble velocity $v_{NL}$ given by Eq.~\ref{eq:goncharov} (dashed-dot green).}
\label{fig:rt_growth}
\end{figure}

\noindent The two approaches differ in how the transition time is defined: \\
a) Matching the transition time at which the bubble velocities match \cite{Layzer}: Following Ref. \cite{Goncharov}, one can define a terminal bubble velocity as 

\begin{equation}
\label{eq:goncharov}
v_{NL} = \sqrt{\frac{2A g}{(1+A) Ck}},
\end{equation}
where C is a constant that depends on dimensionality. Equating this bubble velocity with the derivative of Eq. \ref{eq:rt_piecewise} gives the time for the onset of nonlinearity as

\begin{equation}
t_s = \frac{1}{\gamma} \sinh^{-1} (\frac{1}{\eta_0\gamma} v_{NL})
\end{equation}

\noindent b) Nonlinear effects become important when $k\eta(t_s)  \sim 1$. For the piecewise model shown in Fig.~\ref{fig:rt_growth}, we adopt the threshold \(\eta_s/\lambda=0.1\), corresponding to \(k\eta_s=0.2\pi\approx0.63\). The transition time is then given by Eq. (B7), yielding \(t_s\approx1.86\) ns for the parameters considered. After this transition, we hold the velocity constant at its value at \(t_s\), as given by Eq. (B8).

\begin{equation}
t_s = \frac{\cosh^{-1}(\eta_s/\eta_0)}{\gamma} 
\end{equation}
with the terminal bubble velocity given by 

\begin{equation}
\label{eq:haan}
v_{NL} = \eta_0 \gamma \sinh(\gamma t_s)
\end{equation}

The three solutions (purely the hyperbolic solution, and the growth using the two piecewise methods) are shown in Fig.~\ref{fig:rt_growth}.
\subsection{Summary of model responses}

The same language model evaluations described in the text are used here. Some models identified a transition to nonlinearity. The best responses derived the linear growth rate correctly, selected the $\cosh$ solution or motivated both the exponential and hyperbolic modes, reproduced the amplitude at 1 ns, and warned that the 3.3~ns extrapolation lay outside linear theory but calculated the value incorrectly. A summary of the top four responses is listed in Table~\ref{tab:RT responses}.

The remaining responses fell into two incorrect types. Some used an exponential solution or the hyperbolic solution, identifying answers in the linear regime for both times. A second group made wavelength, acceleration, or time-unit errors, occasionally producing significantly incorrect amplitudes. These results indicate that models are ICF-aware but continue to require expert review.

\begin{table*}[ht]
\centering 
\small
\renewcommand{\arraystretch}{2.0} 

\resizebox{0.8\textwidth}{!}{%
\begin{tabular}{|c|c|} 
\hline
\textbf{Model} & \textbf{Observations} \\ 
\hline
   Claude Sonnet 5 & \makecell{Uses $\cosh$ solution for both times and flags nonlinear behavior \\ but does not calculate the appropriate nonlinear value} \\
\hline
   Inkling Small & \makecell{Explains both choices for the initial mode, flags nonlinear behavior \\ but does not calculate the appropriate nonlinear value} \\
\hline
    GPT-5.4 & Uses exponential solution for both times; ignores nonlinear behavior \\
\hline
   Nemotron Ultra & Uses $\cosh$ solution for both times; ignores nonlinear behavior \\
\hline
\end{tabular}
}
\caption{Observations about the solutions for the top four performing models. The reference solution uses the cosh form, corresponding to an initially stationary interface. None of the models correctly calculates the nonlinear amplitude using either model, although some flag nonlinearity.}
\label{tab:RT responses}
\end{table*}

Five archived retrieval-augmented or frozen-evidence conditions were also examined using GPT-5.6Sol. Retrieval improved discussion of the boundary between linear and nonlinear RT growth relative to the closed evaluations but did not retrieve the answer. This result indicates that additional documents cannot compensate for a modeling assumption that is absent from the prompt. However, this model with its additional reasoning, context, tools, and interactive follow-up in the High thinking mode was able to reverse engineer the solution using the second piecewise bubble velocity model described above. Rephrasing the prompt as ``{\it{The single-mode Rayleigh–Taylor instability transitions from a hyperbolic cosine growth during the linear phase to linear-in-time growth during the nonlinear 
phase. Can you develop a simple model to accurately evaluate the amplitude 
at which this transition occurs? ~\cite{Riccardo} Apply the model to the following problem: A light fluid is supporting a heavy fluid against an acceleration, $g= 8.2 \times 10^{15} cm/s^2$ in the x-direction. The Atwood number A = 0.5. There is an initial sinusoidal perturbation in the y-direction, $\delta x= \eta_0 sin(ky)$. The amplitude, $\eta_0 = 1 \ \mu m$ and the wavelength $\lambda  = 100 \ \mu m$. Note that $k= 2\pi/\lambda$. Write the necessary expressions for the amplitude $\eta$ as a function of time t. Then calculate the numerical value of the amplitude $\eta$ of the perturbation at two times, 1 ns and 3.3 ns.}}" reproduces the correct result using the terminal bubble matching model using the first model described above. However, this prompt does not provide the relevant references motivating the model. 

\section{\label{app:AppC} Capability assessment of LLMs at systematic physicist-like derivation and thinking}
p-$^{11}$B is considered as a favorable option for fusion energy as it produces no product neutrons. Here we present a different example to test the orchestration capability of models. We expect models to rely on their frozen evidence and use a Retrieval Augmented Generation or RAG approach using three references \cite{Sikora, NRLPlasmaFormulary, AtzeniPowerBalance}. 

\noindent Stress test and orchestration study prompt: \\
\indent{\it{Under what conditions does a self-sustaining p–¹¹B plasma state exist?

Using ONLY the supplied references formulate the ignition-relevant equations and implement them in Python. Identify the plasma variables and coupled relations that must be solved self-consistently. Derive the analytic constraints associated with ignition before performing the numerical calculation, and justify all parameter choices and approximations. Ignore external heating. 

Search and optimize over the relevant parameters. Do not prescribe an absolute plasma density to close the calculation. Instead, determine from the equations whether absolute density enters the result, where that dependence originates, and what additional quantities must be specified. Adopt reasonable values and ranges for any slowly varying parameters.

Compare the numerical results with the analytic constraints and investigate any disagreement.

If the nominal model does not produce physical solutions throughout the investigated domain, identify the smallest physically meaningful modification that results in ignition. Quantify the range over which this modification permits admissible solutions and determine how it affects the optimized plasma conditions.

Show the derivation of the analytic model that identifies conditions under which ignition can occur. Generate plots that clearly show the ignition boundary in two different ways (i.e. areal density versus the relevant temperature and the Lawson criterion versus temperature) and the corresponding optimal plasma conditions. Select additional sensitivity plots that reveal which physical assumptions and uncertain inputs most strongly affect the inferred boundary.

Verify the calculated fusion reactivity against the supplied reference. Run the script, inspect its numerical output, correct implementation or convergence problems, and base the conclusions on executed results rather than an untested model. Document the equations, assumptions, numerical methods, optimization procedure, and verification tests.

Deliver the complete derivation with the constraints, Python script, machine-readable results, and figures, together with an assessment of the conditions under which ignition is possible. Treat satisfaction of the analytic ignition criterion as necessary, but not automatically sufficient. What other extensions to this approach can further relax ignition constraints?}}

Note that the prompt does not explicitly inform the model about the specific physics to be extracted from the RAG evidence. The prompt is also deliberately vague about the assumptions associated with the different physics involved in the problem. The appropriate answer must a) derive the ignition criterion from Ref.~\cite{AtzeniPowerBalance}; b) retrieve the relevant equations from Ref.~\cite{NRLPlasmaFormulary}; c) calculate the fusion reactivity from Ref.~\cite{Sikora} and verify it against the figure in the reference; d) derive the power balance equation in the form of a constraint on areal density; e) identify the variables in the problem that can be used to optimize the ignition threshold; f) write a Python script with the appropriate units and debug as needed; g) use an algorithm to optimize the result; h) perform sensitivity analyses as needed. 
\subsection{Analytic solution}
The ignition criterion is given in Ref.~\cite{AtzeniPowerBalance} by 
\begin{equation}
\label{eq:PowerBalance}
    P_{\alpha} - P_{rad} - P_{cond} \geq 0
\end{equation}
where the power deposited by the product $\alpha$ particles, $P_{\alpha}$ must exceed the losses due to radiation, $P_{rad}$ and heat conduction, $P_{cond}$.  
Alpha particle power is given by
\begin{equation}
P_\alpha = f_\alpha n_p n_B Q <\sigma V>, 
\end{equation}
where 
$f_\alpha$ is the fraction of $\alpha$ energy deposited, $n_p$ and $n_B$ are the proton and boron number densities, $Q=8.58$ MeV is the Q-value of the reaction and $<\sigma V>$ is the fusion reactivity. Substituting for $n_p$ and $n_B$ one gets 
\begin{equation}
   \label{eq:fusion}
    P_\alpha = A_f f_\alpha \rho^2 <\sigma V>,
\end{equation}
where $A_f$ is a constant.
The non-relativistic radiation loss is given in Ref.~\cite{NRLPlasmaFormulary} as 
\begin{equation}
P_{rad} \sim T_e^{1/2} n_e \Sigma Z_i^2 n_i,
\end{equation}
where $T_e$ is the electron temperature, $n_e$ is the electron number density, $Z_i$ is the ionization and $n_i$ is the ion density of the protons or boron ions. 
Assuming a fully-ionized p-$^{11}$B plasma and substituting for $n_e$ and $\Sigma Z_i^2 n_i$ one gets
\begin{equation}
\label{eq:radiation}
P_{rad} = A_{Br} T_e^{1/2} \rho^2,
\end{equation}
where $A_{Br}$ is a constant.
One can also derive the energy outflow from the hot spot due to heat conduction as 
\begin{equation}
    P_{cond} = \nabla \cdot q_c \sim - \nabla \cdot \kappa \nabla T_e,
\end{equation}
where the heat conduction coefficient $\kappa \sim A_c T^{5/2}/ln \Lambda $. Substituting, we get
\begin{equation}
\label{eq:HeatConduction}
    P_{cond} = \frac{A_c T_e^{7/2} \rho^2}{ln \Lambda (\rho R)^2}
\end{equation}
Substituting Eqs. ~\ref{eq:fusion}, ~\ref{eq:radiation}, and ~\ref{eq:HeatConduction} into Eq.~\ref{eq:PowerBalance} and rearranging gives
\begin{equation}
\rho R \geq [\frac{\frac{A_c T_e^{7/2}} {\ln \Lambda}} {A_f <\sigma V> f_{\alpha} - A_{Br} T_e^{1/2}}]^{1/2} 
\end{equation}

\subsection{Model results}
\begin{figure}[ht]
\centering
\includegraphics[trim=0cm 0.5cm 0cm 4cm, clip,width=1.0\columnwidth]{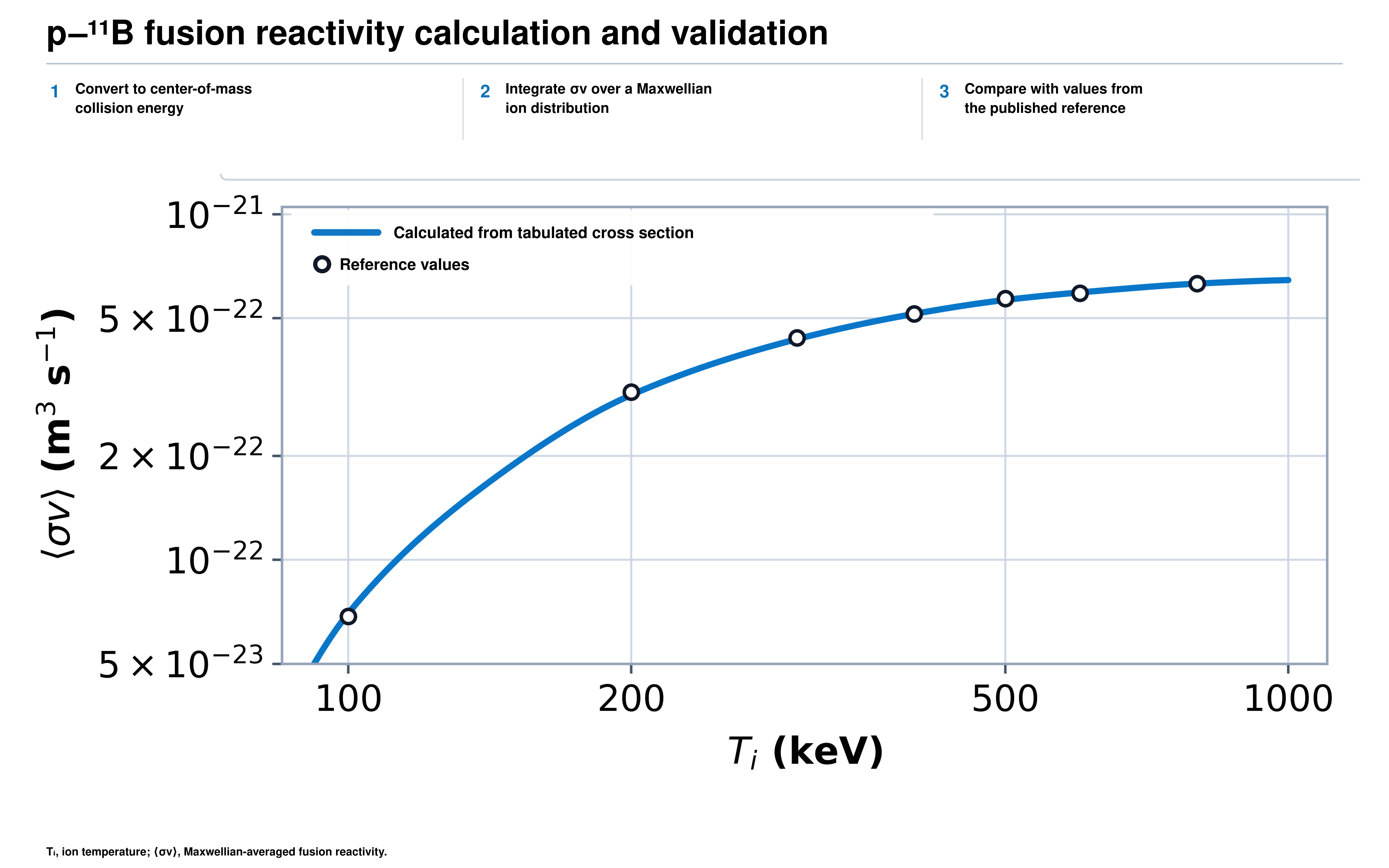}
\caption{Calculated p-$^{11}$B reactivity (line) compared to values obtained by inspection of Fig. 5 in Ref.~\cite{Sikora} versus ion temperature}
\label{fig:pBreactivity}
\end{figure}

The question was tested on GPT-5.6Sol, capable of visually inspecting documents, symbolic derivation and Python scripting, debugging and execution. The result was a dense document along with the Python script and figures, which was read in detail by ICF scientists. The model \\
a) correctly identified the relevant information from each of the RAG references provided. It converted the lab-frame energies for the proton in Tables 1, 2, and 3 to center-of-mass energies and performed integrals over the Maxwellian as described in Eq. 4 of Ref.~\cite{Sikora}, without being explicitly told to do so. Fig.~\ref{fig:pBreactivity} shows this result. The document identifies that the model evaluation of the reactivity gives a maximum discrepancy of 2.44\% from its visual check of the reactivity figure in Ref.~\cite{Sikora}. As part of extensions to the model, the document lists the role of non-Maxwellian distributions on the reactivity. 

\begin{figure}[ht]
\centering
\includegraphics[trim=0cm 0.6cm 0cm 2.7cm, clip,width=0.6\columnwidth]{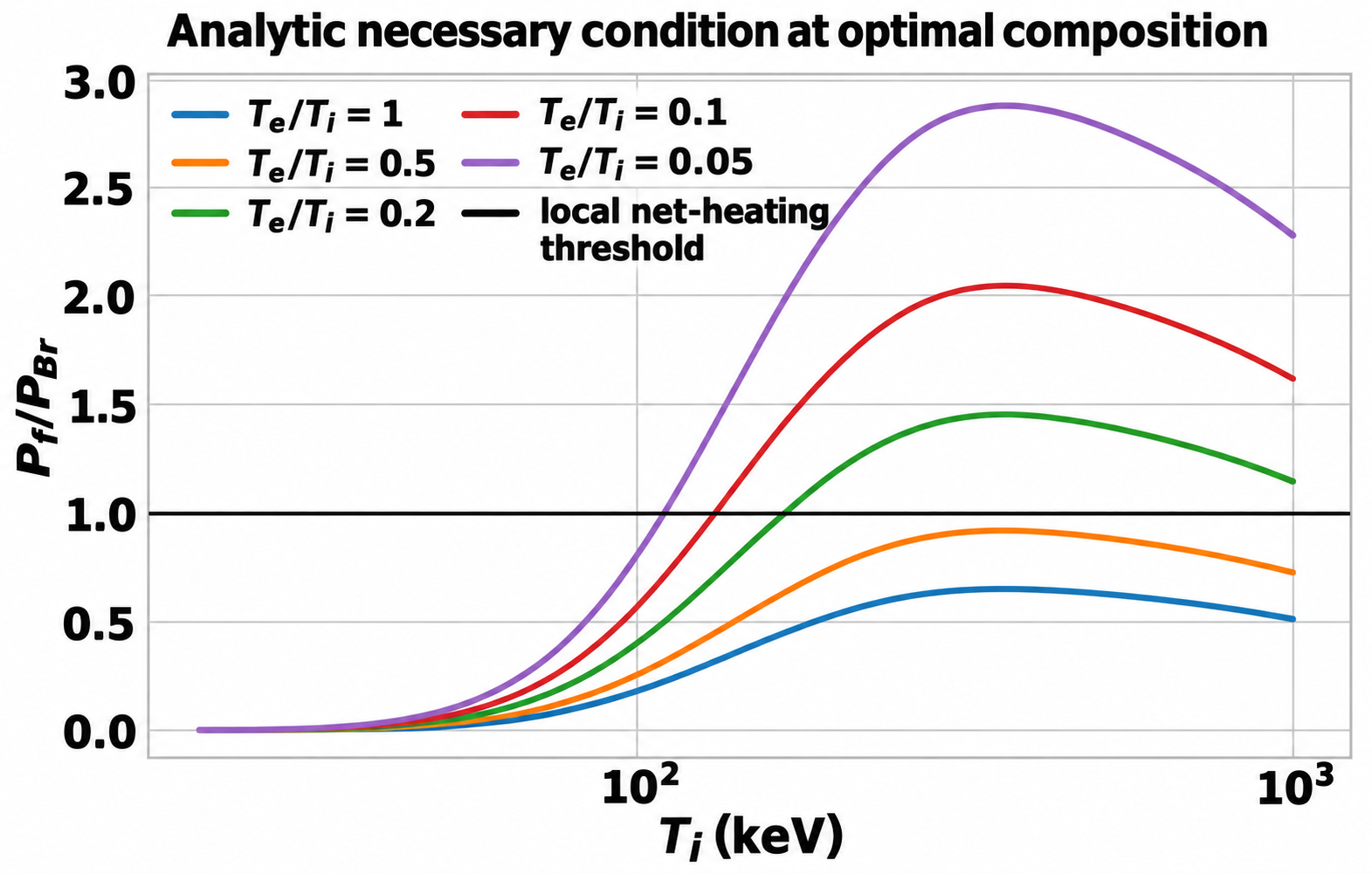}
\caption{Analytic solution for the ratio of the fusion to radiation energy as a function of ion temperature. This is for the analytically derived optimum ratio of boron number fraction. At higher ratios of electron-to-ion temperature no solution is found.}
\label{fig:analyticcondition}
\end{figure}

\noindent b) used the relevant equations from the three documents provided to derive an ignition criterion and identified the variables that it could modify: boron number density fraction, $x_B=n_B/(n_p+n_B)$, two-temperature option, $T_e/T_i$, the fraction of $\alpha$ energy deposited, and the fraction of radiation energy that can be recycled into the system (this is accomplished in the model by a multiplier to the second term in the denominator). The model wrote the equation for the optimal ratio of $x_B$ while correctly identifying that the denominator had to be greater than zero, giving an analytic ratio of 0.0821. For this analytic ratio, the Python script identified where solutions existed for differing ratios of fusion power to radiation losses (Fig.~\ref{fig:analyticcondition}).

\begin{figure}[ht]
\centering
\includegraphics[trim=0cm 0.55cm 0cm 2.7cm, clip,width=0.6\columnwidth]{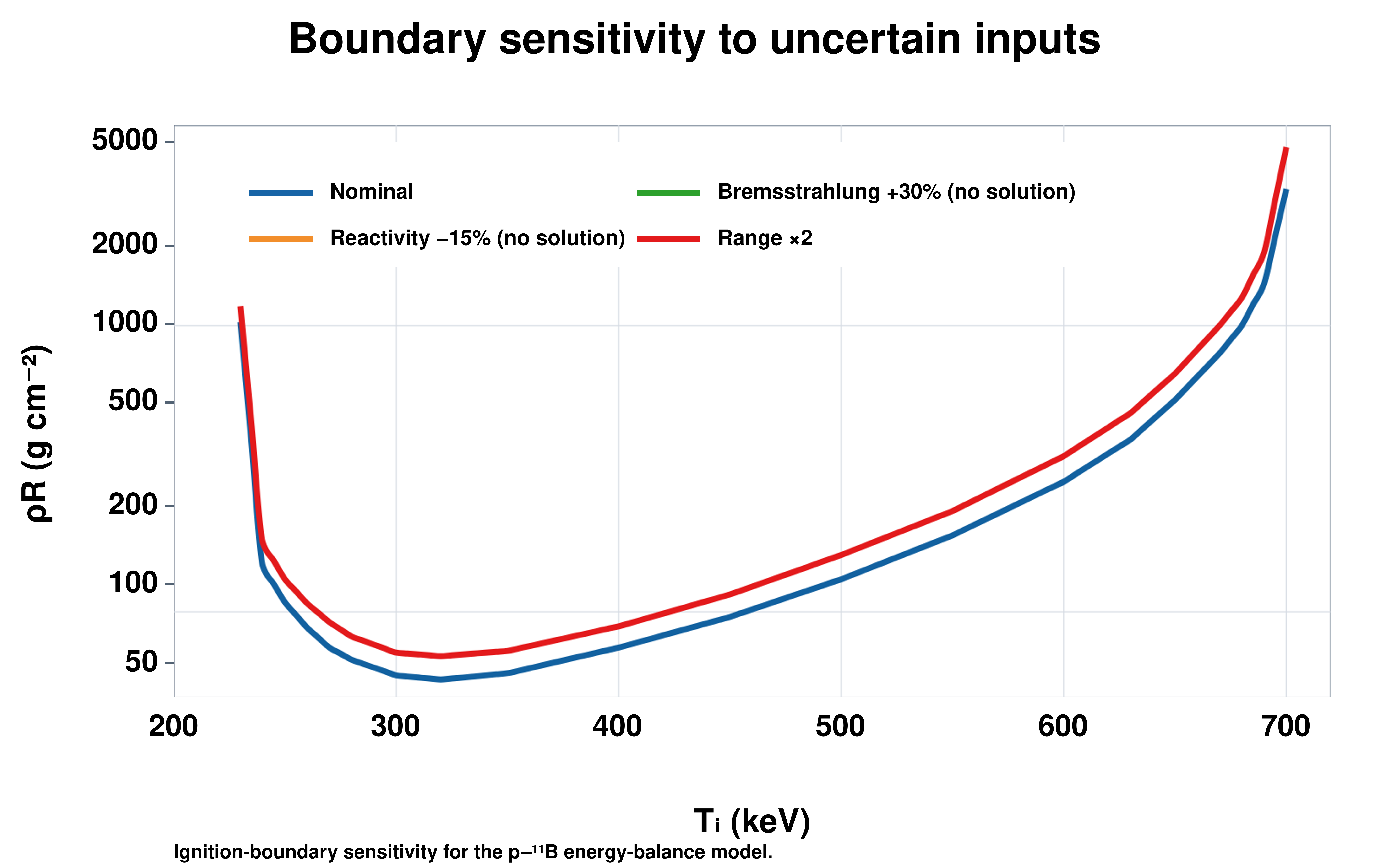}
\caption{Ignition threshold for a hot spot p-$^{11}$B ignition problem. Sensitivity to uncertainties in fusion reactivity, radiation loss, and alpha particle range is also shown. Some lines are missing as they were studied but the model found no plausible physical result.}
\label{fig:IgnitionThreshold}
\end{figure}

\noindent c) pointed out that electrons are relativistic in the temperature regime identified in the problem and applied an overall multiplier to the radiation losses to ``account" for this effect. \\
d) identified that radiation recycling would lower the ignition threshold but did not identify how that could be accomplished. Functionally, the multipliers for radiation losses and recycling are equivalent; the model chooses to treat them separately to illustrate the two concepts.\\
e) correctly pointed out that electron-ion collisions must be accounted for when self-consistently calculating the ion temperature for yields, especially when the ratios deviated significantly from one. \\
f) calculated the ignition threshold conditionally (self-consistent calculations of electron-ion collisions are not included) and performed a sensitivity study on stopping power of the $\alpha$ particles, radiation losses, and fusion reactivity. One ignition threshold formulation is shown in Fig.~\ref{fig:IgnitionThreshold}. ICF scientists checked that these results are consistent with the assumptions provided to the model. 

A summary of plausible scenarios based entirely on the prompt is shown in Table~\ref{tab:pBScenarios}.

\begin{table*}[ht]
\centering 
\renewcommand{\arraystretch}{1.0} 
\resizebox{0.8\textwidth}{!}{%
\begin{tabular}{|l|c|c|c|c|c|c|} 
\hline
\textbf{Scenario} &
\textbf{T$_i$ (keV)} &
\textbf{T$_e$ (keV)} & 
\textbf{x$_B$} & 
\textbf{P$_f$/P$_{Br}$} & 
\textbf{$\rho$R (g/cm$^2$)} &
\textbf{$f_\alpha$} \\ 
\hline
\makecell{Equal temperature \\48.36\% recycled}
        & 286.9 & 286.9 &  0.088 & 1.22 & 82.3 & 0.92  \\
\hline
\makecell{\(T_e/T_i=0.33\) \\No recycling}
        & 320.2 & 106.7 &  0.088 & 1.11 & 58.8 & 0.97 \\
\hline
\makecell{\(T_e/T_i=0.30\) \\No recycling}
        & 306.7 & 92.0 &  0.092 & 1.16 & 36.6 & 0.97 \\
\hline
\makecell{\(T_e/T_i=0.10\) \\No recycling}
        & 243.7 & 24.37 &  0.14 & 1.74 & 4.84 & 0.96 \\
\hline
\end{tabular}
}
\caption{Conditional power balance solutions identified for the p--\(^{11}\mathrm{B}\) ignition model under different temperature and recycling scenarios.}
\label{tab:pBScenarios}
\end{table*}

Based on this study, we conclude that GPT-5.6Sol is plausibly capable of conducting a systematic study involving symbolic derivations, Python scripting and execution, and identifying limiting assumptions where the calculations could be improved. It was clearly evident that the LLM was relying on its training (e.g. conversion from lab to center-of-mass frame for proton energies). The prompt was limited in that it deliberately asked the model to use Ref.~\cite{NRLPlasmaFormulary} in which only nonrelativistic effects are listed, but the model correctly flagged the limitations of the assumptions we expected it to make. The model stayed within the constraints of ``no external heating" and did not identify beam fusion as an option. Though challenging to interpret because of the density of the output, it is critical to have scientists assess these results. We did not further prompt the model to provide it with non-Maxwellian distribution functions or go on to identify radiation losses from relativistic electrons or correct for electron-ion collisions. Using such a tool to delve into the various complex aspects of p-$^{11}$B fusion will be explored elsewhere. This example illustrates the minimal tool set required to assist scientists.

\end{document}